\PassOptionsToPackage{table}{xcolor}
\documentclass{article}

\usepackage{iclr2027_conference,times}

\usepackage{amsmath,amsfonts,bm}

\def\eqref#1{equation~\ref{#1}}

\def\1{\bm{1}}

\DeclareMathAlphabet{\mathsfit}{\encodingdefault}{\sfdefault}{m}{sl}
\SetMathAlphabet{\mathsfit}{bold}{\encodingdefault}{\sfdefault}{bx}{n}

\usepackage{amsmath}
\usepackage{amssymb}

\usepackage{graphicx}
\usepackage{xcolor}

\usepackage{hyperref}
\usepackage{url}
\usepackage{microtype}

\usepackage{booktabs}
\usepackage{tabularx}
\usepackage{multirow}
\usepackage{array}
\usepackage{makecell}
\usepackage{arydshln}

\usepackage{enumitem}

\usepackage{float}
\usepackage{placeins}
\usepackage{needspace}
\usepackage{wrapfig}
\usepackage{capt-of}

\newcolumntype{L}[1]{>{\raggedright\arraybackslash}p{#1}}
\newcolumntype{Y}{>{\raggedright\arraybackslash}X}

\definecolor{oursbg}{RGB}{255,242,242}
\definecolor{groupbg}{RGB}{242,252,252}
\definecolor{avgcol}{RGB}{255,242,242}
\definecolor{softgray}{RGB}{248,248,248}
\definecolor{badbg}{RGB}{252,235,234}
\definecolor{warnbg}{RGB}{255,249,228}

\makeatletter
\newcommand{\shadecell}[1]{%
  \cellcolor{#1}%
  \g@addto@macro\CT@cell@color{\@tempdimb\z@\@tempdimc\z@}%
}
\makeatother

\title{PhyLatent: Learning Dynamics-Relevant Representations for JEPA World Models}

\author{
\textbf{Xi Zeng}\textsuperscript{1,3}
\quad
\textbf{Haojie Ren}\textsuperscript{1,3}
\quad
\textbf{Ziying Song}\textsuperscript{1,2,*}
\\
\textbf{Yuanbo Nie}\textsuperscript{3}
\quad
\textbf{Ross Drummond}\textsuperscript{3}
\\
\textsuperscript{1}School of Mechanical and Aerospace Engineering,
Nanyang Technological University, Singapore
\\
\textsuperscript{2}School of Artificial Intelligence (School of Software),
Yanshan University, Qinhuangdao, China
\\
\textsuperscript{3}The University of Sheffield, Sheffield, United Kingdom
\\
\textsuperscript{*}Corresponding author
\\
\texttt{xzeng991108@gmail.com, rhj130063@gmail.com, songziying@ysu.edu.cn}
}

\iclrfinalcopy

\begin{document}
\maketitle
\lhead{arXiv preprint}

\begin{abstract}
We propose PhyLatent, a dynamics-relevant training objective for
Joint-Embedding Predictive Architecture (JEPA) world models.
Our key observation is that preventing global latent collapse does not
necessarily ensure that the learned representation preserves physically
meaningful state and action relationships.
We identify three failure modes: sensitivity to appearance changes that leave
the physical state unchanged, insufficient separation of distinct physical
states, and insufficient separation of different action-conditioned futures.
We refer to these as Physical Invariance Collapse, Physical Distinguishability
Collapse, and Counterfactual Dynamics Collapse, respectively.
PhyLatent targets these failures through three coordinated training pathways,
implemented with static visual invariance, physical state grounding,
future representation alignment, counterfactual branch separation, and latent denoising.
On OGBench-Cube, PhyLatent reduces the three collapse rates by
\textbf{43.9\%}, \textbf{27.9\%}, and \textbf{47.1\%}, respectively,
while improving model predictive control (MPC) success by \textbf{12.0 percentage points}
(\textbf{17.2\% relative}).
Across four visual-control tasks, average planning success increases by
\textbf{6.62 percentage points} (\textbf{8.3\% relative}).
These results show that global non-collapse alone is insufficient for learning
a reliable JEPA world-model state space, and that explicitly preserving
dynamics-relevant structure can improve closed-loop planning.
\end{abstract}


\section{Introduction}

JEPAs learn compact representations
for future prediction without reconstructing pixel-level details
\citep{lecun2022path}.
Recent work extends this idea to action-conditioned world models, where
predicted latents support control and planning
\citep{assran2025vjepa2,maes2026leworldmodel}.
LeWorldModel (LeWM), for example, combines latent prediction with isotropic
Gaussian regularization to maintain a globally non-degenerate latent
distribution, and performs MPC directly in the learned representation space
\citep{balestriero2025lejepa,maes2026leworldmodel}.

However, we observe a limitation of global anti-collapse regularization:
a globally non-degenerate latent space is not necessarily reliable for
planning.
The latent space may still change with irrelevant visual appearance, fail to
distinguish different physical states, or fail to reflect different action
outcomes.
We examine these failures in Sec.~\ref{sec:three_collapse}.
Specifically, we distinguish:

\begin{itemize}[leftmargin=1.2em,noitemsep,topsep=1pt]
    \item \textbf{Dynamics-relevant structure:}
    physical state information needed for prediction and planning, such as
    robot and object positions and poses, together with how actions change them.

    \item \textbf{Appearance variation:}
    changes in brightness, color, or texture that leave this physical
    information unchanged.
\end{itemize}

This raises a direct question:
\textbf{does global non-collapse preserve dynamics-relevant structure?}

\begin{figure}[!t]
    \centering
    \includegraphics[width=\linewidth]{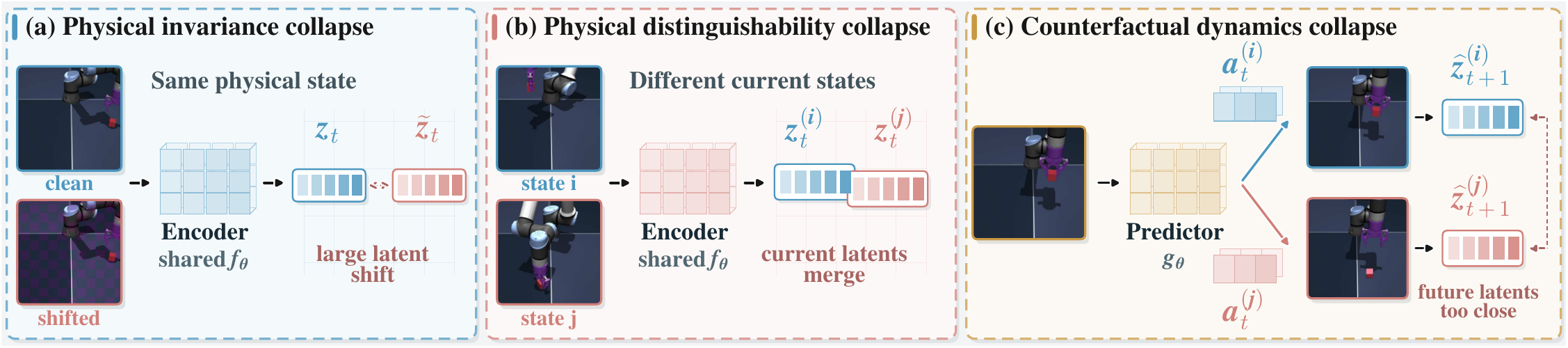}
    \caption{
\textbf{Motivation for PhyLatent.}
    Global non-collapse can coexist with \textbf{(a)} appearance-induced latent
    shifts despite unchanged physical state, \textbf{(b)} insufficient separation
    of distinct physical states, and \textbf{(c)} insufficient predicted separation
    of divergent action-conditioned futures from the same state.
    Here, $z_t$ and $\tilde z_t$ denote clean and appearance-shifted latents,
    $(i),(j)$ index alternative states or action branches, and $\hat z$ denotes a
    predicted future latent.
}
    \label{fig:intro_motivation}
\end{figure}

In LeWM, we identify three forms of dynamics-relevant collapse.
\emph{Physical Invariance Collapse} occurs when changing only visual appearance
breaks a physical relation that was previously represented correctly.
\emph{Physical Distinguishability Collapse} occurs when differences in current
physical states are not sufficiently reflected in latent space.
\emph{Counterfactual Dynamics Collapse} occurs when different actions from the
same current state yield divergent physical outcomes but indistinguishable
predictions from the model.
Figure~\ref{fig:intro_motivation} summarizes the three failures.

To address these failures, we propose \textbf{PhyLatent}, a dynamics-relevant
representation learning method for JEPA world models.
PhyLatent organizes its training objectives into three coordinated pathways.
It suppresses appearance-induced changes, preserves task-relevant physical
information, aligns predicted and observed futures, preserves differences among action-conditioned futures, and regularizes local future structure through latent denoising.
During training, the Physical Grounding pathway uses physical state information
available from simulator trajectories, such as robot and object positions and
poses.
These physical-state targets and all additional auxiliary components are not
required at inference time.

On OGBench-Cube~\citep{ogbench_park2025}, PhyLatent reduces Physical Invariance,
Physical Distinguishability, and Counterfactual Dynamics Collapse from
\textbf{19.58\%}, \textbf{45.51\%}, and \textbf{14.47\%} to
\textbf{10.99\%}, \textbf{32.82\%}, and \textbf{7.65\%}, respectively,
while improving planning success from \textbf{69.67\%} to \textbf{81.67\%}.
Across four tasks, average success increases from \textbf{79.92\%} to
\textbf{86.54\%}.
These results show that global non-collapse is only a starting point:
effective world-model representations benefit from preserving
dynamics-relevant structure.


\section{Related Work}
\label{sec:related_work}

\paragraph{JEPA world models.}
JEPAs learn abstract representations
by predicting target representations rather than reconstructing pixels
\citep{lecun2022path}.
I-JEPA applies joint-embedding prediction to masked image regions
\citep{assran2023ijepa}, building on masked-image modeling paradigms
\citep{he2022mae,bao2022beit}, while V-JEPA extends feature prediction to video \citep{bardes2024vjepa}.
More recent work brings joint-embedding prediction closer to control:
V-JEPA~2 and LeWM introduce action-conditioned prediction for embodied
understanding and planning
\citep{assran2025vjepa2,maes2026leworldmodel}, while PLDM and DINO-WM use
latent dynamics or pretrained visual features for planning
\citep{sobal2025pldm,zhou2025dinowm}.
SD-JEPA instead explicitly structures the latent space by separating
progression from content \citep{thil2026sdjepa}.
Despite these differences, their usefulness for control depends on whether the
learned representation preserves the physical state variables and action
consequences required for planning.

\paragraph{Representation collapse.}
Joint-embedding models require mechanisms that prevent representations from
degenerating into trivial or low-rank embeddings
\citep{grill2020byol,chen2021simsiam,zbontar2021barlow,
bardes2022vicreg,balestriero2025lejepa}.
Existing approaches either use predictive training mechanisms, including
contrastive objectives, asymmetric prediction, momentum targets,
self-distillation, and masked prediction, or directly regularize the
representation distribution through redundancy reduction and
variance--covariance objectives
\citep{oord2018representation,chen2020simclr,he2020moco,
caron2020swav,grill2020byol,chen2021simsiam,caron2021dino,
zbontar2021barlow,bardes2022vicreg,assran2023ijepa,bardes2024vjepa}.
Along the latter direction, LeJEPA introduces Sketched Isotropic Gaussian
Regularization (SIGReg) to maintain a globally non-degenerate embedding
distribution, which LeWM applies to action-conditioned latent prediction
\citep{balestriero2025lejepa,maes2026leworldmodel}.
Sub-JEPA relaxes the full-space Gaussian constraint to multiple random
subspaces \citep{zhao2026subjepa}, while QQWorld further constrains deviations
in the tails of the latent distribution \citep{yu2026qqworld}.
These methods maintain global non-degeneracy, but do not directly constrain
the local structure associated with physical states and action-conditioned
futures.

\paragraph{Dynamics-aware representations and latent control.}
A complementary line of work focuses on representations that retain structure
useful for control, including temporal continuity, controllability,
transition prediction, bisimulation, and inverse dynamics
\citep{jonschkowski2015roboticpriors,sermanet2018tcn,gelada2019deepmdp,
zhang2021bisimulation,wang2022denoisedmdp,pathak2017curiosity}.
For example, Sensorimotor World Models (SMWM) use inverse-dynamics
regularization to learn action-aligned latent representations
\citep{ivashkov2026smwm}.
Learned latent dynamics are also used directly for decision making:
PlaNet performs planning in latent space, Dreamer learns behavior through
latent imagination, and TD-MPC2 performs local trajectory optimization using
a learned latent world model
\citep{hafner2019planet,hafner2020dreamer,hansen2024tdmpc2}.
These works show that latent dynamics can support effective control, but they do not specifically ask whether global non-collapse in JEPA world models preserves the local physical and counterfactual relations required for planning.
PhyLatent targets this representation-level gap.


\begin{figure}[!t]
    \centering
    \includegraphics[width=\linewidth,trim=0 6.4bp 0 2.5bp,clip]{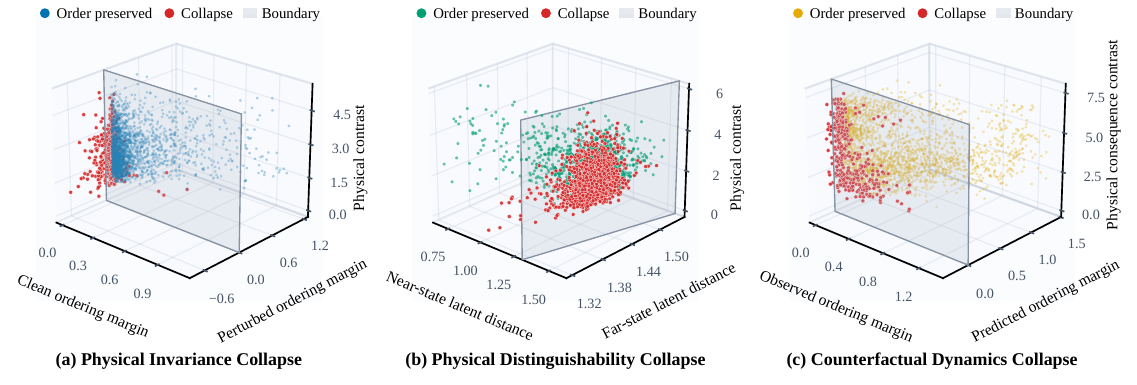}

    \caption{
    \textbf{Sample-level dynamics-relevant collapse on Cube.}
    Each point is an eligible LeWM comparison; red marks ordering violations,
    other colors mark preserved orderings, and translucent planes show their
    boundaries.
    Panels (a)--(c) show Physical Invariance, Physical Distinguishability, and
    Counterfactual Dynamics Collapse, respectively.
    }
    \label{fig:collapse_scatter}
\end{figure}

\section{Why Does LeWorldModel Exhibit Dynamics-Relevant Collapse?}
\label{sec:analysis}
\label{sec:three_collapse}

Figure~\ref{fig:collapse_scatter} shows eligible LeWM comparisons on Cube.
Ordering violations occupy non-negligible regions of all three diagnostic
spaces, indicating that collapse extends beyond isolated rollout examples.

We inspect three representative decisions from failed Cube rollouts of the
same LeWM checkpoint by restoring the simulator at the relevant planning step.
For the Distinguishability and Counterfactual cases, we execute alternative
action sequences from the original cross-entropy method (CEM) candidate pool
from the same state.
Lower latent-space goal cost indicates an outcome judged closer to the goal.
Definitions and replay details appear in
Appendices~\ref{app:diagnostics} and~\ref{app:decision_cases}.

\paragraph{Physical Invariance Collapse.}
Appearance-only perturbations change LeWM's action preferences and lead to a
failed rollout despite unchanged robot, cube, and goal states.
Restoring the original appearance only for the first planning call, from the
same simulator state and with the remaining procedure unchanged, yields success.
Thus, visual appearance alone can change the planning decision.

\paragraph{Physical Distinguishability Collapse.}
The original plan leaves the cube \textbf{10.00 cm} from the goal; an
alternative candidate reaches \textbf{1.49 cm} and succeeds.
Yet directly encoding the realized futures assigns the failed outcome a lower
goal cost (\textbf{207.37 vs.\ 345.24}).
The physically better state is therefore ranked worse by the encoder, locating
the inconsistency in the realized-state representation rather than only in
future prediction.

\paragraph{Counterfactual Dynamics Collapse.}
For the original and alternative plans, directly encoded realized-future costs
(\textbf{227.16 vs.\ 81.32}) correctly favor the alternative, but predicted
costs (\textbf{2.87 vs.\ 55.00}) reverse this ordering and favor the failed plan.
Executing the alternative from the same state succeeds.
Here, the observed inconsistency arises during action-conditioned prediction,
despite correct encoded future ordering.

These interventions locate observed planning inconsistencies without isolating
a unique internal cause.
Together with Figure~\ref{fig:collapse_scatter}, they show that global
non-collapse does not guarantee preservation of local physical and
action-conditioned relations needed for planning.

\begin{figure}[!t]
    \centering
    \includegraphics[width=\linewidth,trim=0 12.9bp 0 9.3bp,clip]{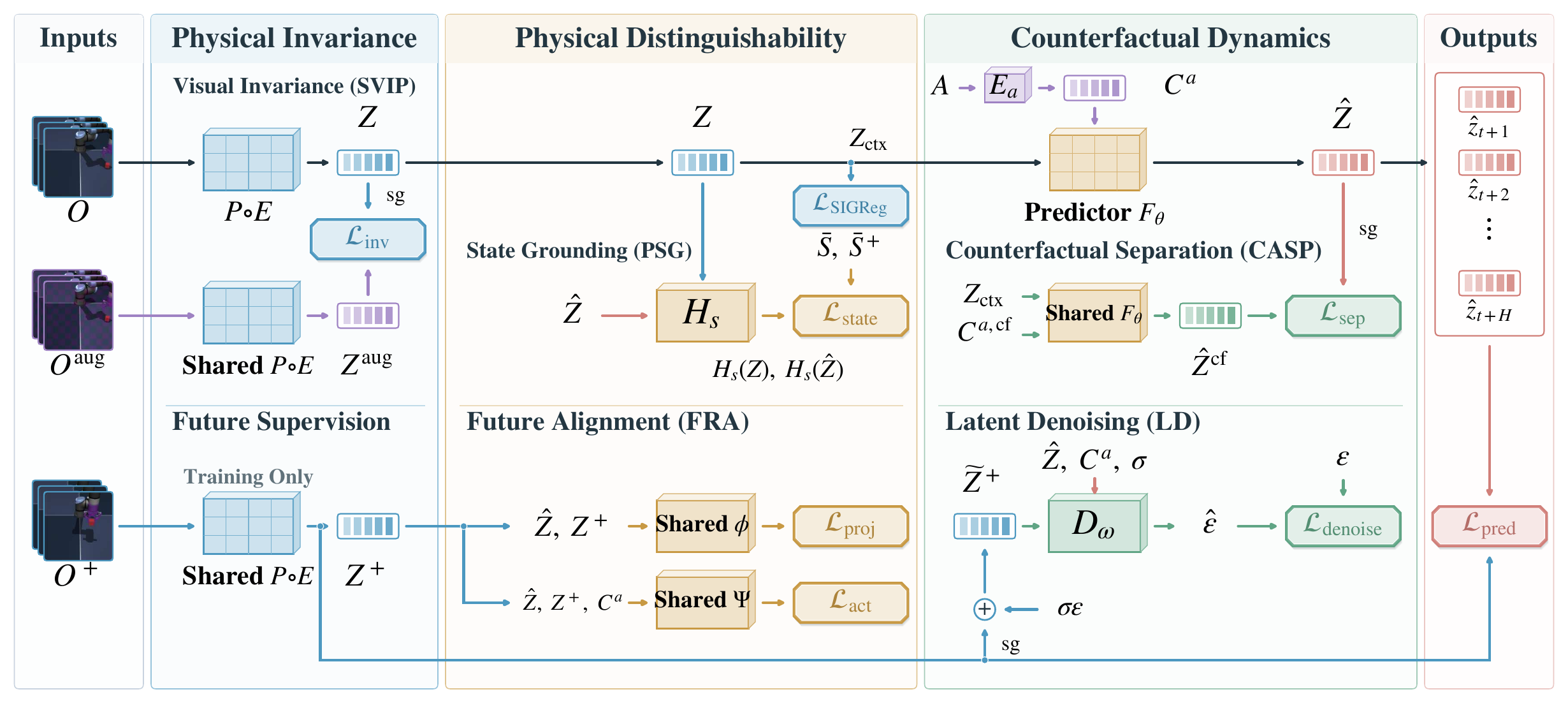}

    \caption{
\textbf{Overview of PhyLatent.}
    Three training-time pathways augment the shared LeWM backbone:
    Physical Invariance (SVIP) constrains appearance consistency;
    Physical Grounding (PSG, FRA) preserves physical-state information and aligns
    predicted with observed futures; Counterfactual Dynamics (CASP, LD) structures
    action-conditioned futures. The original prediction objective is retained.
}
    \label{fig:phylatent_architecture}
\end{figure}

\section{PhyLatent}
\label{sec:method}

Figure~\ref{fig:phylatent_architecture} summarizes the overall PhyLatent
architecture.
PhyLatent retains the LeWM encoder, action encoder, predictor, and
inference-time planning pipeline, while introducing three coordinated
training pathways corresponding to the failure modes identified in
Sec.~\ref{sec:three_collapse}.
Physical Invariance suppresses appearance-induced variation, Physical
Grounding preserves task-relevant state distinctions and aligns future
representations, and Counterfactual Dynamics structures action-conditioned
future predictions.
Although motivated by distinct failure modes, the pathways jointly shape the
same latent representation; all additional heads and supervision are removed
at inference time.

Implementation details, module architectures, sampling procedures, and
task-specific hyperparameters are provided in
Appendix~\ref{app:implementation}.


\subsection{Physical Invariance}
\label{sec:physical_invariance}

For each observation sequence $O$, we construct an appearance-augmented view
$O^{\mathrm{aug}}$ while leaving the underlying physical state unchanged.
The original and augmented observations pass through the same JEPA encoder and
projector, producing $Z$ and $Z^{\mathrm{aug}}$, respectively.
We introduce the \textbf{Static Visual Invariance Penalty (SVIP)}:
\begin{equation}
\mathcal{L}_{\mathrm{inv}}
=
d_{\mathrm N}\!\left(
Z^{\mathrm{aug}},
\operatorname{sg}(Z)
\right).
\label{eq:visual_invariance}
\end{equation}

For feature vectors $X_i,Y_i\in\mathbb{R}^{D}$, we define
\begin{equation}
d_{\mathrm N}(X,Y)
=
\frac{1}{ND}
\sum_{i=1}^{N}
\left\|
\nu(X_i)-\nu(Y_i)
\right\|_2^2,
\qquad
\nu(x)=\frac{x}{\max(\|x\|_2,10^{-12})}.
\label{eq:normalized_feature_distance}
\end{equation}
Here, $N$ counts all non-feature positions, and
$\operatorname{sg}(\cdot)$ denotes stop-gradient, which preserves its forward
value while blocking gradients through the target branch
\citep{chen2021simsiam}.
The detached clean representation serves as the reference, so SVIP penalizes
appearance-induced latent changes without altering the underlying physical
state.


\subsection{Physical Grounding for Distinguishability}
\label{sec:physical_grounding}

\paragraph{Physical State Grounding (PSG).}
When selected physical-state variables are available in the training
trajectories, PSG uses them as auxiliary supervision to ground the latent
representation in task-relevant physical information.
Let $\bar S$ denote the per-dimension standardized physical targets and
$\bar S^{+}$ the corresponding targets temporally aligned with the predicted
future $\hat Z$.
A shared state-prediction head $H_s$ maps both encoded and predicted
representations to these targets using mean squared error (MSE):
\begin{equation}
\mathcal{L}_{\mathrm{state}}
=
\operatorname{MSE}\!\left(H_s(Z),\bar S\right)
+
\operatorname{MSE}\!\left(H_s(\hat Z),\bar S^{+}\right).
\label{eq:state_grounding}
\end{equation}
Sharing $H_s$ across the two branches encourages physical distinctions to
remain recoverable from both encoded and predicted latents.

\paragraph{Future Representation Alignment (FRA).}
FRA extends this grounding to future prediction by aligning $\hat Z$ with the
corresponding encoded future $Z^{+}$.
A shared projection head $\phi$ compares the two representations globally,
whereas an action-query module $\Psi$ summarizes each future relative to the
executed action context.
Let $C^a$ denote the context-aligned action embeddings; their temporal summary
forms the query in $\Psi$, and the future latent sequence provides the keys
and values.
We use
\begin{equation}
\mathcal{L}_{\mathrm{align}}
=
\mathcal{L}_{\mathrm{proj}}
+
\mathcal{L}_{\mathrm{act}},
\label{eq:future_alignment}
\end{equation}
where
$\mathcal{L}_{\mathrm{proj}}
=
d_{\mathrm N}(\phi(\hat Z),\operatorname{sg}(\phi(Z^{+})))$
and
$\mathcal{L}_{\mathrm{act}}
=
d_{\mathrm N}(\Psi(\hat Z,C^a),
\operatorname{sg}(\Psi(Z^{+},C^a)))$.
The two terms align the overall future representation and its
action-conditioned summary, respectively.


\subsection{Counterfactual Dynamics}
\label{sec:counterfactual_dynamics}

\paragraph{Counterfactual Action Separation Penalty (CASP).}
From the same latent context, factual and counterfactual action sequences
produce $\hat Z$ and $\hat Z^{\mathrm{cf}}$ through the shared action encoder
and predictor.
Let $\delta^a_{b,\tau}$ and $\delta^z_{b,\tau}$ denote the normalized action
difference and predicted-future separation for mini-batch sample $b$ and
prediction position $\tau$, respectively.
We define
\begin{equation}
\mathcal{L}_{\mathrm{sep}}
=
\frac{1}{|\Omega|}
\sum_{(b,\tau)\in\Omega}
\left[
\min\!\left(\gamma\delta^a_{b,\tau},m_{\max}\right)
-
\delta^z_{b,\tau}
\right]_+ .
\label{eq:counterfactual_separation}
\end{equation}
Here, $[x]_+=\max(x,0)$, $\Omega$ retains sufficiently different action pairs,
and $\gamma$ and $m_{\max}$ are fixed task-specific hyperparameters.
CASP uses action difference as a proxy for counterfactual branch diversity and
does not explicitly identify outcome-equivalent action pairs.

\paragraph{Latent Denoising (LD).}
LD further regularizes the local neighborhood of the observed future.
We perturb the detached target future as
$\tilde Z^{+}=\operatorname{sg}(Z^{+})+\sigma\epsilon$, where
$\epsilon\sim\mathcal{N}(0,I)$ and $\sigma$ is sampled from a task-specific
noise range.
A denoising head $D_\omega$ receives the noisy target future
$\tilde Z^{+}$, predicted future $\hat Z$, action embedding $C^a$, and noise
scale $\sigma$, and predicts the injected noise:
\begin{equation}
\mathcal{L}_{\mathrm{denoise}}
=
\operatorname{MSE}\!\left(
D_\omega(\tilde Z^{+},\hat Z,C^a,\sigma),
\epsilon
\right).
\label{eq:latent_denoising}
\end{equation}
The detached target provides a fixed reference, and the resulting denoising
signal regularizes local future structure around the observed trajectory.


\subsection{Training Objective}
\label{sec:training_objective}

PhyLatent combines the three pathways with the original LeWM training
objective.
Let $\mathcal{L}_{\mathrm{base}}$ denote the original LeWM objective,
including latent prediction and SIGReg, and
$\mathcal{A}=\{\mathrm{inv,state,align,sep,denoise}\}$ the auxiliary
objectives.
The full objective is
\begin{equation}
\mathcal{L}_{\mathrm{PhyLatent}}
=
\mathcal{L}_{\mathrm{base}}
+
\sum_{i\in\mathcal{A}}
\lambda_i\mathcal{L}_i .
\label{eq:full_objective}
\end{equation}
All auxiliary objectives are optimized jointly with fixed coefficients within
each training run; there is no sequential priority mechanism.
The coefficients are task-specific and were manually adjusted using
preliminary experiments; their values and the remaining training
hyperparameters are reported in Appendix~\ref{app:implementation}.


\section{Experiments}
\label{sec:experiments}

We evaluate PhyLatent on OGBench-Cube \citep{ogbench_park2025},
TwoRooms \citep{sobal2025pldm,maes_lld2026swm,maes2026leworldmodel}, Reacher from the DeepMind Control Suite
\citep{tunyasuvunakool2020}, and PushT \citep{Chi-RSS-23}.
Cube's robot--object geometry supports controlled analysis of all three
failures (Sec.~\ref{sec:analysis}).

\paragraph{Primary evaluation protocol.}

Unless stated otherwise, models share sampled start--goal pairs and simulator
initializations. CEM-MPC uses 300 candidates, 30 elites, prediction and receding
horizons of 5 blocks, an action block size of 5, and a budget of 50 environment
interactions. Cube, TwoRooms, and Reacher use 10 CEM updates; PushT uses 30.
Primary success is reported as mean $\pm$ sample standard deviation across six
paired evaluation seeds. Appendix~\ref{app:implementation} provides full
implementation and evaluation details.


\begin{table}[!t]
\caption{
Closed-loop planning success (\%) across four tasks. PhyLatent, LeWM,
SMWM, and Sub-JEPA use matched paired evaluation (mean $\pm$ across-seed
standard deviation). PLDM and DINO-WM use LeWM's reported baselines
\citep{maes2026leworldmodel}, outside our paired-seed evaluation.
}
\label{tab:world_model_comparison}

\centering
\small
\renewcommand{\arraystretch}{1.03}
\setlength{\tabcolsep}{4.5pt}

\begin{tabular}{
@{}l
!{\color{white}\vrule width 4pt}c
!{\color{white}\vrule width 4pt}c
!{\color{white}\vrule width 4pt}c
!{\color{white}\vrule width 4pt}c
!{\color{white}\vrule width 4pt}c@{}
}
\toprule

\shadecell{groupbg}\textbf{Method}
& \shadecell{groupbg}\textbf{Cube}
& \shadecell{groupbg}\textbf{TwoRooms}
& \shadecell{groupbg}\textbf{Reacher}
& \shadecell{groupbg}\textbf{PushT}
& \shadecell{groupbg}\textbf{Avg.} \\

\midrule

\shadecell{oursbg}\textbf{PhyLatent}
& \shadecell{oursbg}81.67 $\pm$ 3.01
& \shadecell{oursbg}96.83 $\pm$ 2.14
& \shadecell{oursbg}82.17 $\pm$ 3.60
& \shadecell{oursbg}85.50 $\pm$ 3.62
& \shadecell{avgcol}\textbf{86.54} \\

\shadecell{softgray}LeWM
& \shadecell{softgray}69.67 $\pm$ 5.13
& \shadecell{softgray}83.17 $\pm$ 1.47
& \shadecell{softgray}82.33 $\pm$ 2.58
& \shadecell{softgray}84.50 $\pm$ 3.39
& \shadecell{avgcol}79.92 \\

SMWM
& \textbf{83.67 $\pm$ 2.16}
& \textbf{100.00 $\pm$ 0.00}
& 73.17 $\pm$ 2.99
& 85.83 $\pm$ 3.13
& \shadecell{avgcol}85.67 \\

\shadecell{softgray}Sub-JEPA
& \shadecell{softgray}76.83 $\pm$ 3.37
& \shadecell{softgray}94.17 $\pm$ 2.71
& \shadecell{softgray}\textbf{84.83 $\pm$ 2.14}
& \shadecell{softgray}\textbf{89.50 $\pm$ 2.59}
& \shadecell{avgcol}86.33 \\

\midrule

PLDM
& 65.00
& 97.00
& 78.00
& 78.00
& \shadecell{avgcol}79.50 \\

\shadecell{softgray}DINO-WM (w/o proprio.)
& \shadecell{softgray}86.00
& \shadecell{softgray}100.00
& \shadecell{softgray}79.00
& \shadecell{softgray}74.00
& \shadecell{avgcol}84.75 \\

\bottomrule
\end{tabular}

\end{table}

\subsection{Cross-Task Planning Performance}
\label{sec:planning_results}

Table~\ref{tab:world_model_comparison} compares PhyLatent with
LeWM~\citep{maes2026leworldmodel}, SMWM~\citep{ivashkov2026smwm}, and
Sub-JEPA~\citep{zhao2026subjepa} under the same paired evaluation protocol.
PhyLatent raises the four-task average over LeWM from 79.92\% to
\textbf{86.54\%} ($+6.62$ percentage points), with gains of 12.00 and 13.66 points
on Cube and TwoRooms, a modest PushT improvement, and effectively tied Reacher
performance. The backbone and planner are unchanged.

Matched baselines lead on different tasks: SMWM on Cube and TwoRooms, and
Sub-JEPA on Reacher and PushT. PhyLatent's average gain is not uniform
per-task dominance.


\subsection{Dynamics-Relevant Latent Structure}
\label{sec:latent_structure_analysis}

Table~\ref{tab:latent_structure} evaluates the three collapse modes from
Sec.~\ref{sec:three_collapse}. Invariance and Distinguishability collapse decrease
substantially on all four tasks. Counterfactual collapse also decreases in
point estimate, but the Reacher improvement at the planner-aligned horizon
$H{=}5$ is small.

\begin{table}[!t]
\caption{
Collapse rates (\%; lower is better; Appendix~\ref{app:diagnostics}).
CF-H5 denotes Counterfactual Dynamics Collapse at the planner-aligned $H{=}5$.
}
\label{tab:latent_structure}

\centering
\small
\renewcommand{\arraystretch}{1.06}
\setlength{\tabcolsep}{2.4pt}

\begin{tabular}{
@{}l
!{\color{white}\vrule width 4pt}cc
!{\color{white}\vrule width 4pt}cc
!{\color{white}\vrule width 4pt}cc
!{\color{white}\vrule width 4pt}cc@{}
}

\toprule

\shadecell{white}\textbf{Metric}
&
\multicolumn{2}{c}{
\shadecell{groupbg}\textbf{Cube}
}
&
\multicolumn{2}{c}{
\shadecell{groupbg}\textbf{TwoRooms}
}
&
\multicolumn{2}{c}{
\shadecell{groupbg}\textbf{Reacher}
}
&
\multicolumn{2}{c}{
\shadecell{groupbg}\textbf{PushT}
}
\\

\shadecell{white}{}
&
\shadecell{softgray}\textbf{LeWM}
&
\shadecell{oursbg}\textbf{PhyLatent}
&
\shadecell{softgray}\textbf{LeWM}
&
\shadecell{oursbg}\textbf{PhyLatent}
&
\shadecell{softgray}\textbf{LeWM}
&
\shadecell{oursbg}\textbf{PhyLatent}
&
\shadecell{softgray}\textbf{LeWM}
&
\shadecell{oursbg}\textbf{PhyLatent}
\\

\midrule

\textbf{Inv. $\downarrow$}
&
\shadecell{softgray}19.58
&
\shadecell{oursbg}\textbf{10.99}
&
\shadecell{softgray}17.18
&
\shadecell{oursbg}\textbf{4.69}
&
\shadecell{softgray}18.58
&
\shadecell{oursbg}\textbf{7.66}
&
\shadecell{softgray}9.86
&
\shadecell{oursbg}\textbf{3.38}
\\

\textbf{Dist. $\downarrow$}
&
\shadecell{softgray}45.51
&
\shadecell{oursbg}\textbf{32.82}
&
\shadecell{softgray}47.68
&
\shadecell{oursbg}\textbf{26.67}
&
\shadecell{softgray}47.60
&
\shadecell{oursbg}\textbf{33.46}
&
\shadecell{softgray}38.38
&
\shadecell{oursbg}\textbf{30.71}
\\

\textbf{CF-H5 $\downarrow$}
&
\shadecell{softgray}14.47
&
\shadecell{oursbg}\textbf{7.65}
&
\shadecell{softgray}14.10
&
\shadecell{oursbg}\textbf{9.23}
&
\shadecell{softgray}15.48
&
\shadecell{oursbg}15.10
&
\shadecell{softgray}6.63
&
\shadecell{oursbg}\textbf{2.09}
\\

\bottomrule
\end{tabular}

\end{table}

These diagnostics are not the training losses: SVIP, PSG, and CASP optimize
feature consistency, state prediction, and action-based future separation,
whereas the diagnostics test relational ordering under appearance changes,
physical-state differences, and realized action consequences. Sampling and
comparison rules are fixed and shared across models.

Reacher separates improved current-state geometry from improved control:
static diagnostics improve, but planning remains effectively tied. Its
counterfactual gains at shorter horizons do not establish a gain at the
planning horizon, where the paired confidence interval includes zero
(Appendix~\ref{app:paired_diagnostics}).
Held-out rank agreement also improves on fresh random-policy states without
the PSG head or binary collapse definitions
(Table~\ref{tab:app_heldout_geometry}). Absolute correlations remain modest,
supporting improved distance ordering rather than an isometric physical-state
representation or a causal explanation of planning gains.


\subsection{Robustness to Visual Appearance Shifts}
\label{sec:visual_robustness}

\begin{figure}[!htb]
    \centering
    \includegraphics[width=\linewidth,trim=0 2.8bp 0 3.5bp,clip]{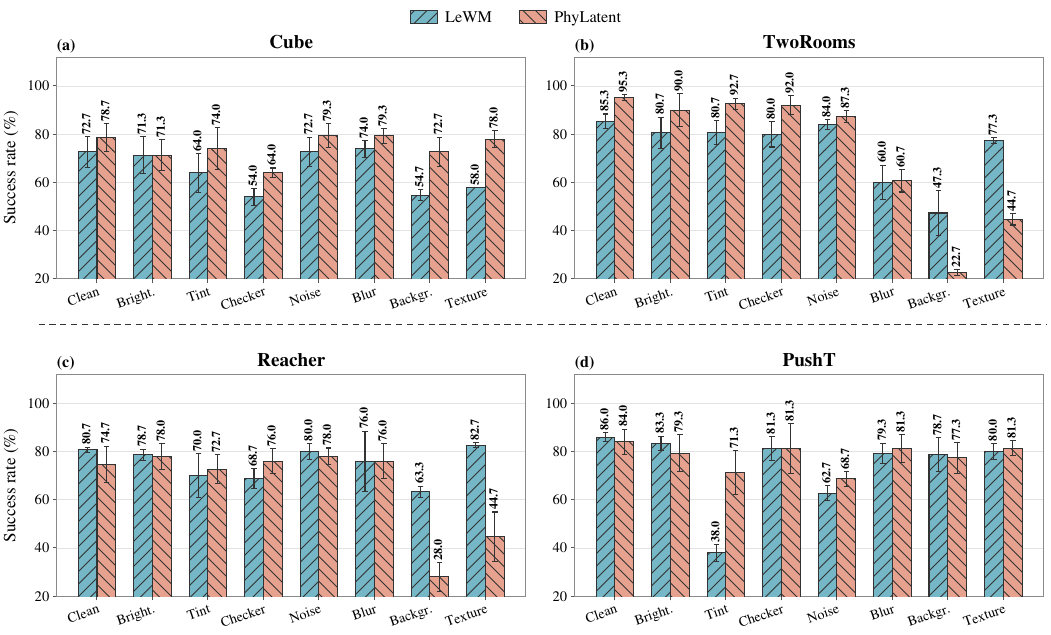}

    \caption{
    \textbf{Closed-loop robustness under visual appearance shifts.}
    Success (\%) of LeWM and PhyLatent under clean observations and controlled
    appearance perturbations across four tasks.
    }

    \label{fig:visual_corruption}
\end{figure}

Figure~\ref{fig:visual_corruption} shows task-dependent robustness. On Cube,
PhyLatent retains advantages of 18.00 and 20.00 percentage points under
background and texture changes. TwoRooms and Reacher instead degrade sharply
under these shifts; TwoRooms grounds only agent position, leaving other scene
information unconstrained. The results support improved invariance of
planning-relevant relations, but not invariance to arbitrary appearance shifts.

\subsection{Supervision and Hyperparameter Controls}
\label{sec:controls}

In Table~\ref{tab:controls_sensitivity}(a), the matched PSG-only control recovers
much of the Cube and TwoRooms improvement but averages 81.75\%, versus 86.54\%
for Full. The PSG-only gains on Cube and TwoRooms coexist with declines
relative to LeWM on Reacher and PushT. Full largely avoids those declines,
indicating a different cross-task profile rather than merely a larger average
gain. This control does not isolate which remaining objective accounts for
that difference. Jointly scaling all five auxiliary losses from
$0.5\times$ to $2\times$ keeps Cube success within 80.17--81.67\% while collapse
rates improve (panel b). The result is stable over this tested weight range,
but better diagnostics do not translate monotonically into better control.

\begin{table}[!htb]
\caption{
Controls for privileged-state supervision and auxiliary-loss weighting.
(a) Mean closed-loop success (\%) for LeWM, its matched PSG-only control, and
PhyLatent across tasks. (b) Joint rescaling of all five PhyLatent auxiliary coefficients on
Cube; success is mean $\pm$ across-seed standard deviation.
}
\label{tab:controls_sensitivity}

\centering
\small
\renewcommand{\arraystretch}{1.05}


\noindent
\begin{minipage}[t]{0.55\linewidth}
\vspace{0pt}
\centering

\setlength{\tabcolsep}{1.20pt}

\begin{tabular}{
@{}l
!{\color{white}\vrule width 2.2pt}c
!{\color{white}\vrule width 2.2pt}c
!{\color{white}\vrule width 2.2pt}c
!{\color{white}\vrule width 2.2pt}c
!{\color{white}\vrule width 2.2pt}c@{}
}

\toprule

\multicolumn{6}{c}{
\shadecell{groupbg}\textbf{(a) Matched privileged-state control}
} \\

\shadecell{white}\textbf{Method}
& \shadecell{white}\textbf{Cube}
& \shadecell{white}\textbf{TwoRooms}
& \shadecell{white}\textbf{Reacher}
& \shadecell{white}\textbf{PushT}
& \shadecell{white}\textbf{Avg.} \\

\midrule

\shadecell{softgray}LeWM
& \shadecell{softgray}69.67
& \shadecell{softgray}83.17
& \shadecell{softgray}\textbf{82.33}
& \shadecell{softgray}84.50
& \shadecell{softgray}79.92 \\

LeWM + PSG
& 80.67
& 96.33
& 67.17
& 82.83
& 81.75 \\

\shadecell{oursbg}\textbf{PhyLatent}
& \shadecell{oursbg}\textbf{81.67}
& \shadecell{oursbg}\textbf{96.83}
& \shadecell{oursbg}82.17
& \shadecell{oursbg}\textbf{85.50}
& \shadecell{oursbg}\textbf{86.54} \\

\bottomrule
\end{tabular}

\end{minipage}%
\hfill%
\begin{minipage}[t]{0.43\linewidth}
\vspace{0pt}
\centering

\setlength{\tabcolsep}{1.30pt}

\begin{tabular}{
@{}c
!{\color{white}\vrule width 2.2pt}c
!{\color{white}\vrule width 2.2pt}c
!{\color{white}\vrule width 2.2pt}c
!{\color{white}\vrule width 2.2pt}c@{}
}

\toprule

\multicolumn{5}{c}{
\shadecell{groupbg}\textbf{(b) Cube auxiliary-weight sensitivity}
} \\

\shadecell{white}\textbf{Scale}
& \shadecell{white}\textbf{Success $\uparrow$}
& \shadecell{white}\textbf{Inv. $\downarrow$}
& \shadecell{white}\textbf{Dist. $\downarrow$}
& \shadecell{white}\textbf{CF-H5 $\downarrow$} \\

\midrule

\shadecell{softgray}$0.5\times$
& \shadecell{softgray}80.17$\pm$3.43
& \shadecell{softgray}12.54
& \shadecell{softgray}34.70
& \shadecell{softgray}9.48 \\

\shadecell{oursbg}$\mathbf{1\times}$
& \shadecell{oursbg}\textbf{81.67$\pm$3.01}
& \shadecell{oursbg}10.99
& \shadecell{oursbg}32.82
& \shadecell{oursbg}7.65 \\

$2\times$
& 80.83$\pm$3.60
& \textbf{8.54}
& \textbf{30.73}
& \textbf{7.55} \\

\bottomrule
\end{tabular}

\end{minipage}

\end{table}


\subsection{Pathway Interactions and Ablations on Cube}
\label{sec:cube_ablation}

Table~\ref{tab:cube_ablation} shows non-additive pathway effects. Removing
Physical Grounding or PSG reduces Cube success from 81.67\% to 76.50\% or
77.00\%, respectively, while distinguishability and counterfactual ordering
also deteriorate. Removing FRA instead lowers Dist. collapse from 32.82\% to
28.85\% while reducing success to 79.33\%. Improving one diagnostic therefore need not improve control.

\begin{table}[!htb]
\caption{
Cube ablations. Non-Full columns remove the named component or group.
Success is mean $\pm$ across-seed standard deviation; lower diagnostics are better.
CF-H5 uses the eligibility intersection across the ablation family
($N{=}10{,}162$), rather than the LeWM--Full pairwise intersection in
Table~\ref{tab:latent_structure}; see Appendix~\ref{app:ablation_success}.
The different intersections explain the different Full CF-H5 values.
}
\label{tab:cube_ablation}

\centering
\small
\renewcommand{\arraystretch}{1.08}
\setlength{\tabcolsep}{2.1pt}

\begin{tabular}{
@{}l
!{\color{white}\vrule width 3pt}c
!{\color{white}\vrule width 3pt}ccc
!{\color{white}\vrule width 3pt}c
!{\color{white}\vrule width 3pt}ccc@{}
}

\toprule

\shadecell{white}{}
&
\shadecell{oursbg}\textbf{Full}
&
\multicolumn{3}{c}{
\shadecell{groupbg}\textbf{Physical Grounding}
}
&
\shadecell{groupbg}\textbf{Invariance}
&
\multicolumn{3}{c}{
\shadecell{groupbg}\textbf{Counterfactual Dynamics}
}
\\

\shadecell{white}\textbf{Metric}
&
\shadecell{oursbg}\textbf{PhyLatent}
&
\shadecell{softgray}\textbf{Group}
&
\textbf{PSG}
&
\textbf{FRA}
&
\shadecell{softgray}\textbf{SVIP}
&
\shadecell{softgray}\textbf{Group}
&
\textbf{CASP}
&
\textbf{LD}
\\

\midrule

\textbf{Success $\uparrow$}
&
\shadecell{oursbg}
\makecell[c]{\textbf{81.67}\\$\pm$3.01}
&
\shadecell{softgray}
\makecell[c]{76.50\\$\pm$3.15}
&
\makecell[c]{77.00\\$\pm$3.63}
&
\makecell[c]{79.33\\$\pm$1.51}
&
\shadecell{softgray}
\makecell[c]{80.67\\$\pm$4.27}
&
\shadecell{softgray}
\makecell[c]{81.17\\$\pm$2.23}
&
\makecell[c]{79.00\\$\pm$0.89}
&
\makecell[c]{79.50\\$\pm$1.97}
\\

\cdashline{1-9}[0.40pt/1.4pt]

\textbf{Inv. $\downarrow$}
&
\shadecell{oursbg}10.99
&
\shadecell{softgray}11.85
&
13.44
&
8.46
&
\shadecell{softgray}9.67
&
\shadecell{softgray}9.74
&
9.46
&
6.93
\\

\textbf{Dist. $\downarrow$}
&
\shadecell{oursbg}32.82
&
\shadecell{softgray}40.11
&
39.33
&
28.85
&
\shadecell{softgray}29.71
&
\shadecell{softgray}27.26
&
28.42
&
28.81
\\

\textbf{CF-H5 $\downarrow$}
&
\shadecell{oursbg}5.934
&
\shadecell{softgray}13.629
&
13.836
&
7.567
&
\shadecell{softgray}8.611
&
\shadecell{softgray}5.737
&
6.908
&
6.288
\\

\bottomrule
\end{tabular}

\end{table}


\subsection{Planning under Increasing Goal Separation}
\label{sec:goal_separation}

\begin{figure}[!htb]
    \centering

    \includegraphics[
        width=\linewidth
    ]{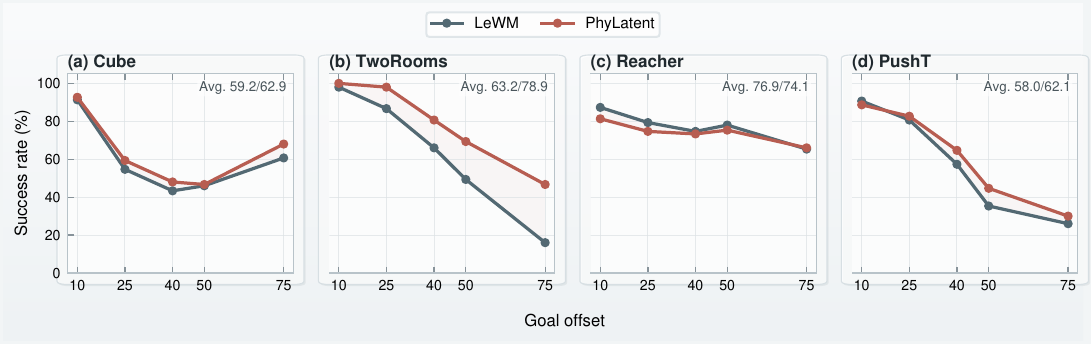}

    \small
    \renewcommand{\arraystretch}{1.00}
    \setlength{\tabcolsep}{2.6pt}

    \begin{tabular}{
    @{}l
    !{\color{white}\vrule width 4pt}c
    !{\color{white}\vrule width 4pt}c
    !{\color{white}\vrule width 4pt}c
    !{\color{white}\vrule width 4pt}c
    !{\color{white}\vrule width 4pt}c@{}
    }

        \toprule

        \multicolumn{6}{c}{
        \shadecell{groupbg}
        \textbf{(e) Cube physical-displacement stratification}
        } \\

        \shadecell{white}\textbf{Method}
        &
        \shadecell{white}
        \makecell[c]{\textbf{Q1}\\\textbf{$<$0.039 cm}}
        &
        \shadecell{white}
        \makecell[c]{\textbf{Q2}\\\textbf{0.039--9.101 cm}}
        &
        \shadecell{white}
        \makecell[c]{\textbf{Q3}\\\textbf{9.101--19.966 cm}}
        &
        \shadecell{white}
        \makecell[c]{\textbf{Q4}\\\textbf{19.966--49.888 cm}}
        &
        \shadecell{white}
        \makecell[c]{\textbf{Q1$\rightarrow$Q4}\\\textbf{Rel. Drop $\downarrow$}}
        \\

        \midrule

        \shadecell{softgray}LeWM
        &
        \shadecell{softgray}\textbf{100.00}
        &
        \shadecell{softgray}81.33
        &
        \shadecell{softgray}48.00
        &
        \shadecell{softgray}26.67
        &
        \shadecell{softgray}73.33
        \\

        \shadecell{oursbg}\textbf{PhyLatent}
        &
        \shadecell{oursbg}\textbf{100.00}
        &
        \shadecell{oursbg}\textbf{88.00}
        &
        \shadecell{oursbg}\textbf{52.00}
        &
        \shadecell{oursbg}\textbf{60.00}
        &
        \shadecell{oursbg}\textbf{40.00}
        \\

        \bottomrule

    \end{tabular}

    \caption{
\textbf{Planning under increasing goal separation.}
(a)--(d) Closed-loop success (\%) at temporal offsets 10, 25, 40, 50, and 75 on
Cube, TwoRooms, Reacher, and PushT, respectively. (e) Cube results by shared
true-displacement quartiles Q1--Q4, with 75 paired instances per quartile.
Larger temporal offsets need not imply monotonically increasing difficulty.
}

    \label{fig:goal_separation}
\end{figure}

\begin{figure}[!t]
    \centering
    \includegraphics[width=0.95\linewidth,trim=0 1.3bp 0 1.1bp,clip]{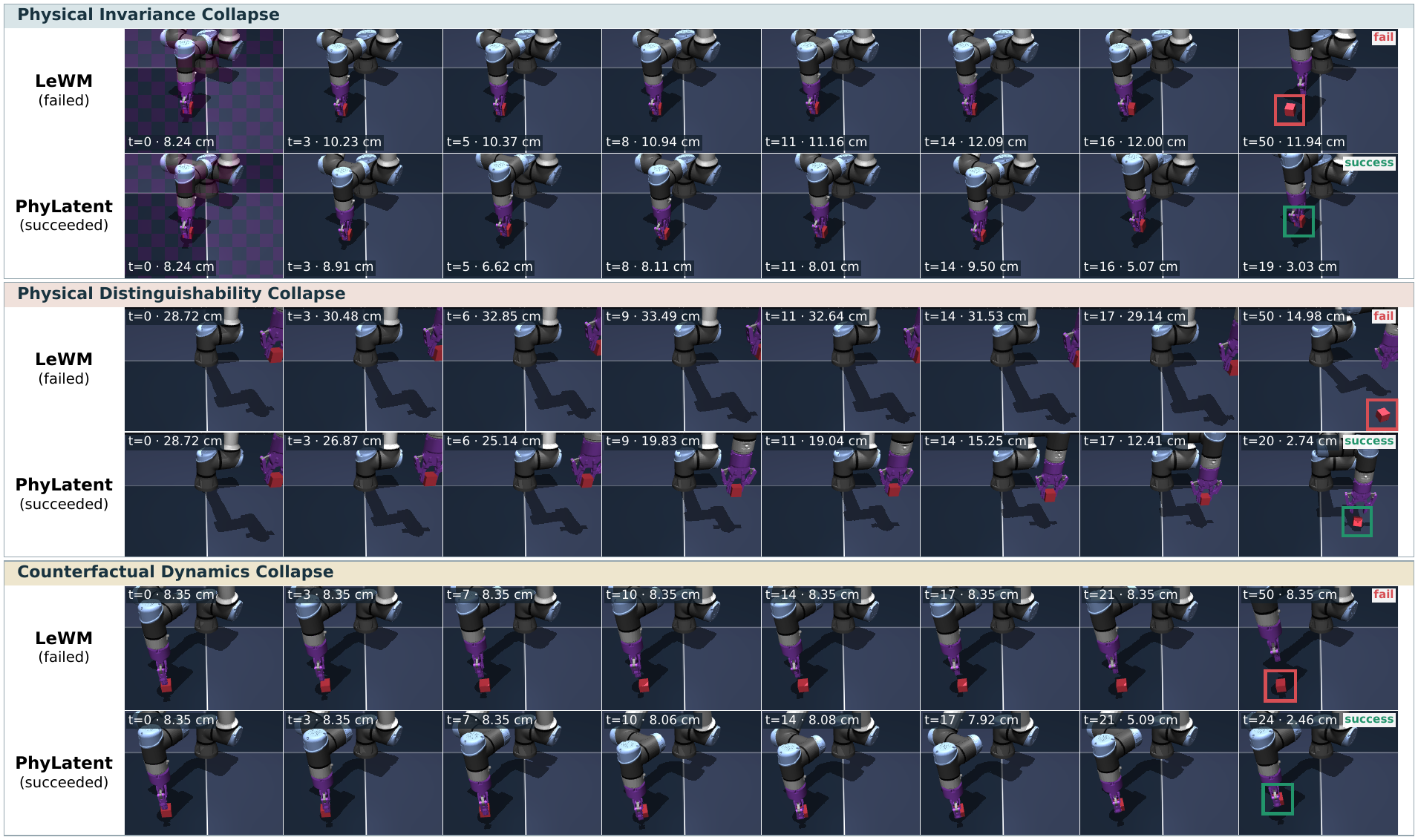}

    \caption{
\textbf{Qualitative rollouts across the three collapse modes.}
Rows compare LeWM and PhyLatent under matched Cube configurations for Physical
Invariance, Physical Distinguishability, and Counterfactual Dynamics Collapse.
}

    \label{fig:qualitative_collapse}
\end{figure}

Figure~\ref{fig:goal_separation} shows higher average success on Cube, TwoRooms,
and PushT, with Reacher the main exception. On TwoRooms, the gap over LeWM grows
from 2.00 points at temporal offset 10 to 30.67 points at offset 75.

Cube provides a physical counterpart: Q1--Q4 are shared quartiles of true
start--goal cube displacement. Both models solve near-zero-motion cases, but
success in the largest quartile rises from 26.67\% with LeWM to 60.00\% with
PhyLatent. 
With temporal offset fixed and bin boundaries shared, the displacement
test compares the models on the same physical separations. Its
largest-quartile advantage shows that the benefit extends beyond
near-stationary goals, although displacement alone does not isolate
every source of planning difficulty.


\subsection{Qualitative Rollout Analysis}
\label{sec:qualitative}

Figure~\ref{fig:qualitative_collapse} compares matched Cube rollouts:
PhyLatent preserves a more useful representation and completes interactions
where LeWM fails.


\section{Conclusion}
\label{sec:conclusion}

PhyLatent exposes a limitation of global anti-collapse objectives in JEPA world
models: a globally non-degenerate representation can still lose the local
physical and action-conditioned relations required for planning.
By targeting Physical Invariance, Physical Distinguishability, and
Counterfactual Dynamics Collapse, PhyLatent reduces the three collapse rates on
Cube from 19.58\%, 45.51\%, and 14.47\% to 10.99\%, 32.82\%, and 7.65\%,
while improving four-task average planning success from 79.92\% to 86.54\%.
The decision-level cases further connect these representation failures to
planning behavior: in the Distinguishability case, an alternative candidate
reaches 1.49\,cm from the goal instead of 10.00\,cm when the latent ordering
fails.
Together, these results show that preserving dynamics-relevant structure can
improve both latent representations and closed-loop planning.

\paragraph{Limitation and Future Work.}
PhyLatent uses selected physical-state variables during training, and its
auxiliary objectives are not fully orthogonal in the shared representation
space.
Future work should reduce this supervision requirement and better understand
and coordinate the interactions among the three training pathways.


\subsection*{AI use statement}

In this work, we used generative AI tools for retrieval and discovery of
relevant literature, to draft portions of the manuscript, and to aid and
polish the writing.
All AI-assisted material was reviewed and revised by the authors, who take
full responsibility for the final content of the paper.


\subsection*{Reproducibility statement}

We provide the methodological and evaluation details needed to reproduce
PhyLatent in the main paper and appendices.
Appendix~\ref{app:diagnostics} defines the collapse diagnostics and their
sampling protocol; Appendix~\ref{app:decision_cases} documents the
decision-level case studies; and Appendix~\ref{app:implementation} provides
the implementation, training, hyperparameter, and evaluation settings.
Additional diagnostic analyses, robustness and planning results, and full
ablation results are reported in
Appendices~\ref{app:diagnostic_results},
\ref{app:robustness_planning}, and
\ref{app:ablation_full}, respectively.


\bibliography{iclr2027_conference}
\bibliographystyle{iclr2027_conference}


\clearpage
\appendix


\setcounter{table}{0}
\renewcommand{\thetable}{A\arabic{table}}

\setcounter{figure}{0}
\renewcommand{\thefigure}{A\arabic{figure}}


\section{Collapse Diagnostic Definitions}
\label{app:diagnostics}

We define the three diagnostics using task-relevant physical relations and
their corresponding latent distances.
Physical bins, references, appearance perturbations, action branches, and
random seeds are fixed before model evaluation and shared across models.


\subsection{Physical Adapters Used by the Diagnostics}
\label{app:diagnostic_adapters}

Training-time PSG targets and evaluation-time physical adapters are distinct.
Angles in the diagnostic vectors are represented as
$\varphi(\theta)=[\sin\theta,\cos\theta]$ to avoid the artificial
$-\pi/\pi$ discontinuity.

\begin{table}[!t]
\centering
\caption{
\textbf{Task-specific physical adapters used by the collapse diagnostics.}
The frame-level vector defines physical distances for Invariance and
Distinguishability, while the reset state is used to generate counterfactual
futures from the same simulator state.
}
\label{tab:app_physical_adapters}

\small
\renewcommand{\arraystretch}{1.06}
\setlength{\tabcolsep}{3.5pt}

\begin{tabular}{@{}
>{\raggedright\arraybackslash}p{0.12\linewidth}
>{\raggedright\arraybackslash}p{0.52\linewidth}
>{\raggedright\arraybackslash}p{0.24\linewidth}
@{}}
\toprule

\rowcolor{groupbg}[0pt][0pt]
\multicolumn{3}{c}{
\textbf{Diagnostic physical representations}
} \\

\rowcolor{softgray}[0pt][0pt]
\textbf{Task}
& \textbf{Frame-level physical vector}
& \textbf{CF reset state} \\

\midrule

Cube &
End-effector position; end-effector yaw (sin/cos);
gripper opening/contact; block position; block yaw (sin/cos).
& Full simulator $qpos/qvel$. \\

\cdashline{1-3}[0.40pt/1.4pt]

PushT &
Agent $(x,y)$; block $(x,y)$; block angle (sin/cos).
& Full PushT state. \\

\cdashline{1-3}[0.40pt/1.4pt]

Reacher &
Joint configuration (sin/cos); fingertip position; target position.
& $qpos$ and $qvel$. \\

\cdashline{1-3}[0.40pt/1.4pt]

TwoRooms &
Agent position $(x,y)$.
& Agent position and target state. \\

\bottomrule
\end{tabular}
\end{table}

For Reacher, the frame-level diagnostic intentionally omits velocity because
it measures goal-relevant configuration geometry; counterfactual restoration
still includes $qvel$.
For TwoRooms, the frame-level metric should be interpreted as agent-position
geometry rather than a complete Markov-state metric.


\subsection{Diagnostic Sampling Protocol}
\label{app:diagnostic_protocol}

Table~\ref{tab:app_diagnostic_protocol} summarizes the sampling settings.

\begin{table}[!t]
\centering
\caption{
\textbf{Sampling protocol for the collapse diagnostics.}
All sampling choices are fixed before model evaluation and shared across
LeWM and PhyLatent.
}
\label{tab:app_diagnostic_protocol}

\small
\renewcommand{\arraystretch}{1.07}
\setlength{\tabcolsep}{3.5pt}

\begin{tabular}{@{}
>{\raggedright\arraybackslash}p{0.23\linewidth}
>{\raggedleft\arraybackslash}p{0.13\linewidth}
>{\raggedright\arraybackslash}p{0.30\linewidth}
>{\raggedleft\arraybackslash}p{0.16\linewidth}
@{}}
\toprule

\rowcolor{groupbg}[0pt][0pt]
\multicolumn{4}{c}{
\textbf{Collapse-diagnostic sampling}
} \\

\rowcolor{softgray}[0pt][0pt]
\textbf{Parameter}
& \textbf{Value}
& \textbf{Parameter}
& \textbf{Value} \\

\midrule

Diagnostic seeds
& 4100--4102
& Frame anchors / seed
& 1,000 \\

Reference pool
& 4,096
& References / anchor / physical bin
& 64 \\

CF anchors / seed
& 300
& Action branches / anchor
& 7 (21 pairs) \\

CF horizons
& $H\in\{1,3,5\}$
& Hierarchical bootstrap
& 10,000 \\

\bottomrule
\end{tabular}
\end{table}

\paragraph{Reference sampling.}
Each diagnostic seed uses 1,000 frame anchors and a pool of 4,096 reference
frames. We sample 64 references per physical-distance quartile, allowing
replacement when a bin is too small. Invariance concatenates 64 position-wise
$Q_1$--$Q_3$ comparisons and 64 $Q_1$--$Q_4$ comparisons per anchor;
Distinguishability uses the 64 $Q_1$--$Q_4$ comparisons. These are paired
reference lists, not their Cartesian product.

\paragraph{Paired hierarchical bootstrap.}
Each of 10,000 replicates samples three seeds with replacement from
4100--4102, then resamples anchors with replacement within each sampled seed:
1,000 for Invariance/Distinguishability and 300 for CF. Compared models share
the same resampled indices. All references, perturbation conditions, and
contrasts associated with an anchor remain grouped; eligibility masks and
physical high/low selections are fixed before resampling.
For CF, each replicate pools the resampled failure and eligible counts before
computing the two models' rates and their difference; it does not average
anchor-level rate ratios.
The 95\% interval comprises the 2.5th and 97.5th percentiles of finite paired
differences, using linear interpolation of quantiles. Non-finite replicates
are excluded. Differences are PhyLatent minus LeWM in
Table~\ref{tab:app_diagnostic_stats} and ablation minus Full in
Table~\ref{tab:app_cube_ablation_ci}, in percentage points.
Bootstrap RNG seeds are 92000 for Invariance, 92001 for Distinguishability,
and $93000+H$ for the primary CF comparisons.


\subsection{Common Latent and Physical Distances}

For observation $x$,
\begin{equation}
\tilde z(x)
=
\frac{z(x)}
{\max(\lVert z(x)\rVert_2,\varepsilon)},
\qquad
d_z(x_i,x_j)
=
\left\lVert
\tilde z(x_i)-\tilde z(x_j)
\right\rVert_2 .
\label{eq:app_latent_norm}
\end{equation}

Physical coordinates are robust-standardized using the interquartile range
(IQR) and shared dataset statistics:
\begin{equation}
\tilde s_k
=
\frac{
s_k-\operatorname{median}(s_k)
}{
\operatorname{IQR}(s_k)/1.349+\varepsilon
},
\qquad
d_{\mathrm{phy}}(i,j)
=
\left\lVert
\tilde s_i-\tilde s_j
\right\rVert_2 .
\label{eq:app_physical_distance}
\end{equation}
Here, $\varepsilon>0$ is a small numerical stabilizer.


\subsection{Physical Invariance Collapse}
\label{app:invariance_diagnostic}

For an observation $x$, let $n$ and $f$ denote physically near and far
reference observations.
We first require the clean latent relation to be correct.
For an appearance-only transform $T$,
\begin{equation}
E^{\mathrm{Inv}}
=
\mathbb{1}[d_z(x,n)<d_z(x,f)],
\qquad
F^{\mathrm{Inv}}
=
\mathbb{1}[d_z(Tx,Tn)\ge d_z(Tx,Tf)] .
\label{eq:app_inv_eligibility}
\end{equation}

For perturbation condition $c$,
\begin{equation}
C_{\mathrm{Inv}}^{(c)}
=
100
\frac{
\sum E^{\mathrm{Inv}}F^{\mathrm{Inv}}
}{
\sum E^{\mathrm{Inv}}
},
\qquad
C_{\mathrm{Inv}}
=
\frac{1}{9}
\sum_{c=1}^{9}
C_{\mathrm{Inv}}^{(c)} .
\label{eq:app_inv_rate}
\end{equation}

The nine conditions comprise three severity levels each for brightness,
tint, and checker perturbations.
Eligibility coverage is
\[
\mathrm{Coverage}_{\mathrm{Inv}}
=
100
\frac{
\sum E^{\mathrm{Inv}}
}{
\#\text{candidate triplets}
}.
\]

\paragraph{Diagnostic appearance transforms.}
Let $x=I/255$ for an RGB uint8 image $I$ and let
$\ell\in\{0.5,1,1.5\}$ denote severity. Before clipping, brightness applies
$(1-0.16\ell)x+0.07\ell$, tint applies
$x\odot(1+0.16\ell,1-0.12\ell,1+0.05\ell)$, and checker applies
$x+\ell M(u,v)(0.08,-0.03,0.06)$, where
\[
M(u,v)=\left(\left\lfloor u/c_{\rm cell}\right\rfloor+
\left\lfloor v/c_{\rm cell}\right\rfloor\right)\bmod 2,
\qquad
c_{\rm cell}=\max\!\left(1,\operatorname{round}
\frac{\min(H_{\rm img},W_{\rm img})}{12}\right).
\]
Here $H_{\rm img},W_{\rm img}$ are image dimensions; a $224\times224$ image
uses 19-pixel cells. Each deterministic transform is shared by the anchor and
all its references. Transformed pixels are multiplied by 255, clipped to
$[0,255]$, cast to uint8, divided by 255, and ImageNet-normalized with RGB
mean $(0.485,0.456,0.406)$ and standard deviation $(0.229,0.224,0.225)$.
This diagnostic path performs no resizing. The encoder/projector outputs are
then unit-normalized as in Eq.~\ref{eq:app_latent_norm}.
The checker resolution here differs from the fixed 18-pixel cells used in the
closed-loop stress test in Appendix~\ref{app:visual_stress}.


\subsection{Physical Distinguishability Collapse}
\label{app:distinguishability_diagnostic}

For each observation $x$, near and far references are sampled according to
the shared physical-distance distribution, with $n\in Q_1$ and $f\in Q_4$
in the primary definition:
\begin{equation}
F^{\mathrm{Dist}}
=
\mathbb{1}[d_z(x,f)\le d_z(x,n)],
\qquad
C_{\mathrm{Dist}}
=
100
\frac{
\sum F^{\mathrm{Dist}}
}{
\#\text{sampled near/far comparisons}
}.
\label{eq:app_dist_rate}
\end{equation}

Equality, NaN, and Inf are counted as failures.
Alternative physical pairings are reported only as sensitivity analyses.


\subsection{Counterfactual Dynamics Collapse}
\label{app:counterfactual_diagnostic}

\paragraph{Action-branch calibration and anchor selection.}
The diagnostic branches are distinct from the stochastic CASP training pairs.
For each task, we take up to 200,000 equally spaced dataset actions, subtract
the coordinate-wise median $m$, and divide by $s=\operatorname{IQR}/1.349$.
Coordinates with $s\le10^{-8}$ use $\max(\mathrm{std},10^{-6})$ instead;
NaN standardized values are replaced by zero. Principal component analysis of
the standardized-action covariance supplies three directions $v_k$; for
two-dimensional actions, the third is the normalized sum of the two principal
directions. If $r$ is the 75th percentile of standardized action norms, the
seven branches use
\[
a_0=\operatorname{clip}_{[-1,1]}(m),\qquad
a_{k,\pm}=\operatorname{clip}_{[-1,1]}(m\pm r\,s\odot v_k),
\quad k=1,2,3.
\]
The radii are 4.178787288, 0.896791558, 1.318597874, and 1.942412591 for
Cube, TwoRooms, Reacher, and PushT, respectively.
Each action is held constant for 25 environment steps; $H=1,3,5$ therefore
correspond to 5, 15, and 25 steps. The calibrated actions are fixed across
anchors, seeds, and models within a task.

For diagnostic seed $s_{\rm eval}$, RNG seed $s_{\rm eval}+100000$ samples
up to 900 candidate valid indices without replacement. We retain the first
300 for which all seven branches complete 25 steps without early termination
or truncation. Each branch restores the same simulator state and reset seed.
The model context uses observations at $t-10,t-5,t$ and five-action blocks
standardized by their separately computed block means and standard deviations.

For each current-state anchor, seven action branches produce
$\binom{7}{2}=21$ branch pairs at horizons $H\in\{1,3,5\}$.
For each pair $p$, we record the true future physical distance
$D_{\mathrm{phy}}$, encoded true-future distance
$D_{\mathrm{obs},m}$, and predicted-future distance
$D_{\mathrm{pred},m}$.

Within each anchor, pairs are ranked by $D_{\mathrm{phy}}$.
The bottom five form the low-consequence set and the top five the
high-consequence set, yielding up to 25 high-vs.-low contrasts;
exact physical ties are excluded.

For a high-consequence pair $p^+$ and low-consequence pair $p^-$,
\begin{equation}
\begin{aligned}
E_m^{\mathrm{CF}}
&=
\mathbb{1}
[
D_{\mathrm{obs},m}(p^+)
>
D_{\mathrm{obs},m}(p^-)
],
\\
F_m^{\mathrm{CF}}
&=
\mathbb{1}
[
D_{\mathrm{pred},m}(p^+)
\le
D_{\mathrm{pred},m}(p^-)
].
\end{aligned}
\label{eq:app_cf_eligibility}
\end{equation}

Formal model comparison uses the same eligible contrasts:
\begin{equation}
E_{\mathrm{common}}
=
E_{\mathrm{LeWM}}^{\mathrm{CF}}
\land
E_{\mathrm{PhyLatent}}^{\mathrm{CF}},
\qquad
C_{\mathrm{CF},m,\mathrm{matched}}^{(h)}
=
100
\frac{
\sum E_{\mathrm{common}}F_m^{\mathrm{CF}}
}{
\sum E_{\mathrm{common}}
}.
\label{eq:app_cf_matched}
\end{equation}

$H{=}5$ matches the planner horizon and is the primary result;
$H{=}1$ and $H{=}3$ are sensitivity analyses.



\section{Decision-Level Failure Cases}
\label{app:decision_cases}

This section provides additional details for the three representative
planning failures discussed in Sec.~\ref{sec:three_collapse}.
All cases are taken from failed closed-loop Cube rollouts using the same
LeWM checkpoint.
For each case, we restore the simulator state at the planning step where
the decision was made and inspect candidate futures evaluated by the planner.
The interventions are used only for case-level analysis and do not modify
or retrain the world model.

We distinguish two quantities.
The \emph{predicted goal cost} is assigned before execution to the future
produced by the action-conditioned predictor.
The \emph{encoded realized-future cost} is obtained after executing a
candidate in the simulator and directly encoding the resulting future
observation.
Lower values indicate outcomes judged by the latent-space planner to be
closer to the goal.

\paragraph{Planner cost and replay endpoint.}
Let $z(o)=P(E(o))$ and $z_g=z(o_g)$. The reported costs use the planner's
terminal objective:
\[
J_{\rm pred}(A)=\|\hat z_{t+25}(A)-z_g\|_2^2,
\qquad
J_{\rm obs}(A)=\|z(o^A_{t+25})-z_g\|_2^2.
\]
These projected embeddings are not unit-normalized. The squared differences
are summed across features, without division by the feature dimension or
summation across time. Thus these costs differ from the normalized geometric
diagnostic distances in Appendix~\ref{app:diagnostics}.
Candidate replays retain the common 25-step endpoint for cost comparison even
when the success region is reached earlier; reported success times are the
first entry into that region.

\subsection{Physical Invariance Case}

The case begins from dataset frame 20164 at rollout step $t=0$.
The simulator state, target, LeWM checkpoint, and planning randomness are
held fixed.
In the original rollout, the appearance-perturbed planning input leads to
failure within the 50-step evaluation budget.

For the intervention, only the current and goal observations supplied to
the first planning call are restored to their original appearance.
The simulator state and subsequent closed-loop planning procedure are
otherwise unchanged.
With this single change, the rollout reaches the Cube success threshold
at environment step 21.

This intervention isolates the visual input at the decision point:
the physical decision problem is unchanged, but the planning outcome changes
with appearance.

\subsection{Physical Distinguishability Case}

The case begins from dataset frame 82020 at rollout step $t=0$.
The original CEM search already evaluated an alternative action sequence
at iteration 9, candidate 208 (zero-based indexing).

When the original and alternative action sequences are executed from the
same simulator state, the original plan leaves the cube 10.00 cm from the
goal, whereas the alternative reaches 1.49 cm and enters the success region
at environment step 23.
Their complete physical-state distances are 2.026 and 0.473, respectively.

Despite this physical ordering, directly encoding the two realized futures
gives a goal cost of 207.37 for the failed outcome and 345.24 for the
successful alternative.
The encoded representation therefore ranks the physically worse state as
closer to the goal.

\subsection{Counterfactual Dynamics Case}

The case begins from dataset frame 891917 at rollout step $t=0$.
The alternative corresponds to iteration 3, candidate 112
(zero-based indexing) of the original CEM search.

Directly encoding the realized futures gives goal costs of 227.16 and 81.32
for the original and alternative plans, respectively, correctly favoring
the alternative.
Before execution, however, the action-conditioned predictor assigns
predicted goal costs of 2.87 and 55.00, reversing this ordering.
Executing the alternative from the same simulator state reaches the
success region at environment step 22, whereas the original rollout does
not succeed within the evaluation budget.

\subsection{Scope of the Case Analysis}

The interventions above identify where an inconsistency in the planning
relation can be observed, but they do not isolate a unique causal mechanism
inside the model.
In the Distinguishability and Counterfactual cases, the intervention replaces
the executed action sequence rather than modifying an individual
representation module.

CEM does not directly execute the single lowest-cost discrete candidate.
The reported alternatives were evaluated during the original search,
whereas the final CEM action is obtained through the elite-distribution
update.
The replays should therefore be interpreted as controlled decision-level
counterfactuals rather than exact reconstructions of an alternative planner
output.



\section{Implementation and Evaluation Details}
\label{app:implementation}


\subsection{Notation}
\label{app:notation}

Table~\ref{tab:app_notation} summarizes symbols used repeatedly in PhyLatent;
local indices and dimensions are defined where they first appear.

\begin{table}[!t]
\centering
\caption{\textbf{Notation used in PhyLatent.}}
\label{tab:app_notation}
\small
\renewcommand{\arraystretch}{0.98}
\setlength{\tabcolsep}{2.1pt}

\begin{tabular}{@{}
L{0.083\linewidth}L{0.207\linewidth}
!{\color{white}\vrule width 3pt}
L{0.083\linewidth}L{0.207\linewidth}
!{\color{white}\vrule width 3pt}
L{0.083\linewidth}L{0.207\linewidth}
@{}}
\toprule
\shadecell{groupbg}\textbf{Symbol} & \shadecell{groupbg}\textbf{Meaning}
& \shadecell{groupbg}\textbf{Symbol} & \shadecell{groupbg}\textbf{Meaning}
& \shadecell{groupbg}\textbf{Symbol} & \shadecell{groupbg}\textbf{Meaning} \\
\midrule

$O$ & Input observation sequence.
& $O^{\mathrm{aug}}$ & Appearance-augmented observations.
& $O^{+}$ & Target-future observations. \\

$A$ & Factual action sequence.
& $A^{\mathrm{cf}}$ & Counterfactual actions.
& $E$ & Shared visual encoder. \\

$P$ & JEPA projector.
& $E_a$ & Shared action encoder.
& $F_\theta$ & Action-conditioned predictor. \\

$Z$ & $P(E(O))$, encoded latent sequence.
& $Z_{\mathrm{ctx}}$ & Predictor latent context.
& $Z^{+}$ & Encoded target future. \\

$\hat Z$ & Predicted factual future.
& $\hat Z^{\mathrm{cf}}$ & Predicted counterfactual future.
& $Z^{\mathrm{aug}}$ & Augmented-view latent. \\

$C^a$ & Context-aligned action embeddings.
& $C^{a,\mathrm{cf}}$ & Counterfactual action embeddings.
& $H_s$ & PSG state head. \\

$\bar S,\bar S^{+}$ & Standardized physical targets.
& $\phi$ & FRA projection head.
& $\Psi$ & FRA action-query module. \\

$\delta^a,\delta^z$ & CASP action/future gaps.
& $\gamma,m_{\max}$ & CASP margin scale/cap.
& $\Omega$ & Active CASP pair set. \\

$D_\omega$ & LD denoising head.
& $\tilde Z^{+}$ & Noisy detached target future.
& $\epsilon,\sigma$ & LD noise and scale. \\

$\operatorname{sg}$ & Stop-gradient.
& $d_{\mathrm N}$ & Normalized feature distance.
& $\lambda_i$ & Auxiliary loss weight. \\

\makecell[c]{$\mathcal{L}_{\mathrm{base}}$} & Original LeWM prediction + SIGReg.
& \makecell[c]{$H_s(Z)$\\$H_s(\hat Z)$} & Physical-state predictions.
& & \\

\bottomrule
\end{tabular}
\end{table}


\subsection{Training-Time Auxiliary Modules}
\label{app:training_details}

PhyLatent retains the LeWM visual backbone, projector, action encoder,
action-conditioned predictor, and inference-time MPC planner.
The latent dimension is 192.
Training clips contain four observations:
$Z=[z_1,z_2,z_3,z_4]$.
The first three positions form the predictor context, while
$Z^{+}=[z_2,z_3,z_4]$ is the temporally aligned target future.

\paragraph{Training data and clip construction.}
The Cube and Reacher datasets each contain 10,000 episodes, 2,010,000 frame
rows, and 1,820,000 valid clips. TwoRooms contains 10,000 episodes, 920,809
frame rows, and 730,809 clips; PushT contains 18,685 episodes, 2,336,736 frame
rows, and 1,981,721 clips. The source files are
\texttt{cube\_single\_expert.h5}, \texttt{reacher.h5},
\texttt{tworoom.h5}, and \texttt{pusht\_expert\_train.h5}, respectively.
Images are $224\times224$ RGB uint8 arrays. Each sliding-window clip contains
observations at offsets 0, 5, 10, and 15 and twenty low-level actions.
Starts advance by one dataset step, giving
$\sum_e\max(L_e-19,0)$ clips for episode lengths $L_e$.

Training and validation use an approximately 90/10 random split of clips,
with split/loader seed 3072. Neighboring and overlapping clips from the same
episode can therefore occur in both subsets. Images are converted to floating
point, ImageNet-normalized, and then resized to $224\times224$.
Non-image training columns are standardized per dimension using all non-NaN
rows before this split. 


\paragraph{SVIP appearance augmentation.}
SVIP modifies only the input pixels; the corresponding actions,
physical-state targets, and temporal indices remain unchanged.
For an ImageNet-normalized input frame $x$,
\begin{equation}
x^{\mathrm{aug}}_{b,t,c,h,w}
=
x_{b,t,c,h,w}
+
u_{b,t}
+
v_{b,t,c},
\qquad
u_{b,t}\sim\mathcal U(-\beta,\beta),
\quad
v_{b,t,c}\sim\mathcal U(-\chi,\chi).
\label{eq:app_svip_augmentation}
\end{equation}

The brightness offset is shared across channels and spatial locations within
one frame, whereas the channel offset is shared spatially.
Both are sampled independently for each sample and temporal position.


\paragraph{PSG targets and standardization.}
Each physical dimension is standardized using dataset-level statistics:
\begin{equation}
\mu_k
=
\frac{1}{N}\sum_{n=1}^{N}s_{n,k},
\qquad
\sigma_k
=
\sqrt{
\frac{1}{N-1}
\sum_{n=1}^{N}
(s_{n,k}-\mu_k)^2
},
\qquad
\bar s_{n,k}
=
\frac{s_{n,k}-\mu_k}{\sigma_k}.
\label{eq:app_psg_standardization}
\end{equation}

For $S=[s_1,s_2,s_3,s_4]$, the prediction-aligned targets are
$\bar S^{+}=[\bar s_2,\bar s_3,\bar s_4]$.


\begin{table}[!t]
\centering
\caption{
\textbf{Training-time implementation details.}
Panel (a) lists the physical targets used by PSG.
Panel (b) summarizes the training-only auxiliary heads.
}
\label{tab:app_training_details}

\small
\renewcommand{\arraystretch}{1.05}


\setlength{\tabcolsep}{4pt}

\begin{tabular}{@{}
l
c
p{0.70\linewidth}
@{}}
\toprule

\rowcolor{groupbg}[0pt][0pt]
\multicolumn{3}{c}{
\textbf{(a) Task-specific PSG targets}
} \\

\rowcolor{softgray}[0pt][0pt]
\textbf{Task}
& \textbf{Dim.}
& \textbf{Physical targets} \\

\midrule

Cube & 28 &
Arm joint positions and velocities; end-effector position and yaw
cosine/sine features; gripper opening and contact; cube position,
quaternion, and yaw cosine/sine features. \\

\cdashline{1-3}[0.40pt/1.4pt]

Reacher & 6 &
Joint positions, finger-to-target displacement, and joint velocities. \\

\cdashline{1-3}[0.40pt/1.4pt]

PushT & 7 &
Agent position, block position and angle, and agent velocity. \\

\cdashline{1-3}[0.40pt/1.4pt]

TwoRooms & 2 &
Agent position $(x,y)$; target and door states are excluded. \\

\bottomrule
\end{tabular}

\vspace{0.55em}


\setlength{\tabcolsep}{3.2pt}

\begin{tabular}{@{}
c
p{0.64\linewidth}
c
c
@{}}
\toprule

\rowcolor{groupbg}[0pt][0pt]
\multicolumn{4}{c}{
\textbf{(b) Auxiliary-module architectures}
} \\

\rowcolor{softgray}[0pt][0pt]
\textbf{Module}
& \textbf{Architecture}
& \textbf{Out.}
& \textbf{Params.} \\

\midrule

$H_s$ &
$\mathrm{LN}(192)\rightarrow
\mathrm{Linear}(192,512)\rightarrow
\mathrm{GELU}\rightarrow
\mathrm{Linear}(512,d_s)$
& $d_s$
& task-dep. \\

\cdashline{1-4}[0.40pt/1.4pt]

$\phi$ &
$\mathrm{LN}(192)\rightarrow
\mathrm{Linear}(192,128)\rightarrow
\mathrm{GELU}\rightarrow
\mathrm{Linear}(128,128)$
& 128
& 41,600 \\

\cdashline{1-4}[0.40pt/1.4pt]

$\Psi$ &
Action summary as query; future latents as keys/values;
four-head attention followed by $\mathrm{LN}(192)$
& 192
& 259,776 \\

\cdashline{1-4}[0.40pt/1.4pt]

$D_\omega$ &
$\mathrm{LN}(577)\rightarrow
\mathrm{Linear}(577,768)\rightarrow
\mathrm{GELU}\rightarrow
\mathrm{Linear}(768,768)\rightarrow
\mathrm{GELU}\rightarrow
\mathrm{Linear}(768,192)$
& 192
& 1,183,298 \\

\bottomrule
\end{tabular}

\end{table}

LN, MHA, and GELU denote layer normalization, multi-head attention, and
Gaussian error linear unit, respectively.


\paragraph{FRA modules.}
Let $\bar A$ denote the per-dimension standardized action blocks.
The same $\phi$ is applied to predicted and target futures, with
stop-gradient applied to the complete target-side output $\phi(Z^{+})$.

Five consecutive low-level actions are concatenated into one standardized
action block and encoded by the shared action encoder:
\[
C^a
=
E_a(\bar A_{\mathrm{ctx}})
\in
\mathbb{R}^{B\times3\times192}.
\]

For $\Psi$, the action embeddings are averaged over the three temporal
positions,
\[
\bar c_b^a
=
\frac{1}{3}
\sum_{\tau=1}^{3}
C^a_{b,\tau},
\]
and used as a single query:
\begin{equation}
\Psi(X,C^a)
=
\operatorname{LN}
\left[
\operatorname{MHA}
\left(
W_q\bar c^a+b_q,\,
W_kX+b_k,\,
W_vX+b_v
\right)
\right].
\label{eq:app_action_query}
\end{equation}

The future latent sequence provides the keys and values.
We use four attention heads with 48 dimensions per head.


\paragraph{CASP sampling and distances.}
Counterfactual actions are generated using a mini-batch permutation $\pi$,
shared across prediction positions, followed by Gaussian perturbation:
\begin{equation}
\bar A^{\mathrm{cf}}_{b,\tau}
=
\bar A_{\pi(b),\tau}
+
\eta_{b,\tau},
\qquad
\eta_{b,\tau}
\sim
\mathcal N(0,\sigma_a^2I).
\label{eq:app_casp_sampling}
\end{equation}

The normalized gaps are
\begin{equation}
\delta^a_{b,\tau}
=
\frac{
\|
\bar A^{\mathrm{cf}}_{b,\tau}
-
\bar A_{b,\tau}
\|_2
}{
\sqrt{D_a}
},
\qquad
\delta^z_{b,\tau}
=
\frac{
\|
\hat Z^{\mathrm{cf}}_{b,\tau}
-
\operatorname{sg}(\hat Z_{b,\tau})
\|_2
}{
\sqrt{D_z}
}.
\label{eq:app_casp_distances}
\end{equation}

Here, $D_a$ denotes the standardized action-block dimension and
$D_z=192$ is the latent dimension.
The active set is
\begin{equation}
\Omega
=
\left\{
(b,\tau):
\delta^a_{b,\tau}
>
\operatorname{median}_{b,\tau}(\delta^a)
\right\}.
\label{eq:app_casp_active}
\end{equation}

The median is computed over all batch--prediction-position action gaps in
the current mini-batch.
If the active set is empty, all pairs are used as a fallback.
CASP does not explicitly identify actions that lead to physically equivalent outcomes.


\paragraph{LD noise and denoising head.}
For each sample and prediction position,
\[
\sigma_{b,\tau}
\sim
\mathcal U(\sigma_{\min},\sigma_{\max})
\]
independently.
The scale has shape $B\times3\times1$ and is broadcast over the latent
dimension.

The denoiser input is
\begin{equation}
U
=
\operatorname{Concat}
[
\tilde Z^{+},
\hat Z,
C^a,
\sigma
]
\in
\mathbb{R}^{B\times3\times577},
\qquad
\hat\epsilon
=
D_\omega(U).
\label{eq:app_ld_input}
\end{equation}

The 577 dimensions correspond to $192+192+192+1$.
$D_\omega$ predicts the injected noise rather than the clean latent.
Only $Z^{+}$ is detached when forming the noisy target.


\subsection{Task-Specific Auxiliary Hyperparameters}
\label{app:aux_hyperparameters}

\begin{table}[!t]
\centering
\caption{
\textbf{Task-specific auxiliary hyperparameters.}
Objective weights are shown on the left; augmentation, separation, and
denoising parameters are shown on the right.
}
\label{tab:app_aux_hyperparameters}

\footnotesize
\renewcommand{\arraystretch}{1.10}

\begin{minipage}[t]{0.47\linewidth}
\centering
\setlength{\tabcolsep}{1.45pt}
\begin{tabular}{@{}lcccccc@{}}
\toprule
\rowcolor{groupbg}[0pt][0pt]
\multicolumn{7}{c}{\textbf{(a) Objective weights}} \\
\rowcolor{softgray}[0pt][0pt]
\textbf{Task}
& $\lambda_{\rm inv}$
& $\lambda_{\rm state}$
& $\lambda_{\rm align}$
& $\lambda_{\rm sep}$
& $\lambda_{\rm den}$
& $\lambda_{\rm sig}$ \\
\midrule
Cube     & .050 & .250 & .100 & .0200 & .0100 & .090 \\

TwoRooms & .030 & .200 & .080 & .0200 & .0100 & .090 \\
Reacher  & .003 & .045 & .025 & .0015 & .0005 & .090 \\

PushT    & .025 & .150 & .060 & .0180 & .0080 & .090 \\
\bottomrule
\end{tabular}
\end{minipage}
\hspace{0.035\linewidth}
\begin{minipage}[t]{0.47\linewidth}
\centering
\setlength{\tabcolsep}{1.45pt}
\begin{tabular}{@{}lcccccc@{}}
\toprule
\rowcolor{groupbg}[0pt][0pt]
\multicolumn{7}{c}{\textbf{(b) Augmentation, CASP, and LD}} \\
\rowcolor{softgray}[0pt][0pt]
\textbf{Task}
& $\beta$
& $\chi$
& $\gamma$
& $m_{\max}$
& $\sigma_a$
& $\sigma_{\rm LD}$ \\
\midrule
Cube     & .040 & .030 & .080 & 1.00 & .100 & $[.03,.35]$ \\

TwoRooms & .030 & .020 & .080 & 1.00 & .100 & $[.05,.35]$ \\
Reacher  & .012 & .008 & .025 & .35 & .025 & $[.05,.10]$ \\

PushT    & .030 & .020 & .080 & 1.00 & .080 & $[.05,.30]$ \\
\bottomrule
\end{tabular}
\end{minipage}
\end{table}

The coefficients are fixed within each training run and were manually adjusted
using observations from preliminary experiments.
We do not claim a globally optimal weighting or a uniform exhaustive
hyperparameter search.


\subsection{Optimization Details}
\label{app:optimization}

The original LeWM prediction objective has unit weight and SIGReg uses
$\lambda_{\mathrm{sigreg}}=0.09$.
We use AdamW~\citep{loshchilov2019adamw} with learning rate $5\times10^{-5}$ and weight decay
$10^{-3}$, together with gradient clipping at $1.0$.

No auxiliary-objective warm-up, coefficient schedule, dynamic loss
weighting, or sequential optimization is used.
All enabled objectives are combined into a single scalar loss and optimized
jointly from the beginning of training.



\subsection{Primary Evaluation Protocol}
\label{app:primary_protocol}

MPC~\citep{mayne2000mpc} uses CEM~\citep{deboer2005cem} for action-sequence search.
Closed-loop comparisons use paired start--goal instances and simulator
initializations across compared models. Table~\ref{tab:app_primary_protocol}
summarizes the protocol underlying Table~\ref{tab:world_model_comparison}.

For the paired comparisons, each method uses one fixed checkpoint per task.
Seeds 600--605 measure evaluation variability, not variability across six
independent training runs. LeWM and Sub-JEPA use released task-specific
checkpoints; SMWM uses a model trained from scratch for each task.
Shared planning settings do not imply identical training budgets or auxiliary
supervision across methods.

\begin{table}[!t]
\centering
\caption{\textbf{Primary closed-loop evaluation protocol.}
Reported $\pm$ values denote sample standard deviation across six paired
evaluation seeds.}
\label{tab:app_primary_protocol}
\small
\renewcommand{\arraystretch}{1.01}

\begin{minipage}[t]{0.50\linewidth}
\vspace{0pt}
\centering
\setlength{\tabcolsep}{3.0pt}
\begin{tabular}{@{}lr@{}}
\toprule
\multicolumn{2}{c}{\shadecell{groupbg}\textbf{(a) MPC / CEM setup}} \\
\shadecell{white}\textbf{Parameter} & \shadecell{white}\textbf{Value} \\
\midrule
Goal offset & 25 dataset steps \\
Evaluation budget & 50 env. steps \\
Planning / receding horizon & 5 / 5 blocks \\
Action block & 5 env. actions \\
CEM samples / top-$k$ & 300 / 30 \\
CEM iterations & 10; PushT: 30 \\
Evaluation seeds & 600--605 \\
Episodes / seed / model / task & 100 \\
Episodes / task / model & 600 \\
\bottomrule
\end{tabular}
\end{minipage}
\hspace{0.035\linewidth}
\begin{minipage}[t]{0.45\linewidth}
\vspace{0pt}
\centering
\setlength{\tabcolsep}{2.2pt}
\begin{tabular}{@{}l p{0.60\linewidth}@{}}
\toprule
\multicolumn{2}{c}{\shadecell{groupbg}\textbf{(b) Native success conditions}} \\
\shadecell{white}\textbf{Task} & \shadecell{white}\textbf{Condition} \\
\midrule
Cube & 3D cube-goal error $\le 0.04$ m. \\
TwoRooms & Agent-goal 2D pixel distance $<16$. \\
Reacher & $\lvert q_j-q_j^{\rm goal}\rvert<0.05$ rad for every joint. \\
PushT & $\lVert s_{0:4}-s^{\rm goal}_{0:4}\rVert_2<20$ and periodic angular error $<\pi/9$. \\
\bottomrule
\end{tabular}
\end{minipage}
\end{table}

The horizon is expressed in action blocks: five blocks of five low-level
actions correspond to 25 environment steps. An episode is successful if the
native task condition is reached at any time within the 50-step budget.

\subsection{Control Experiment Definitions}
\label{app:control_definitions}

The matched PSG control in Table~\ref{tab:controls_sensitivity} adds the same
PSG objective and task-specific physical targets used by PhyLatent to LeWM,
while disabling SVIP, FRA, CASP, and LD. The auxiliary-weight sweep keeps the
LeWM objective and SIGReg unchanged and jointly scales the five PhyLatent
auxiliary coefficients by $0.5\times$, $1\times$, or $2\times$; no coefficient
is retuned independently within the sweep.


\section{Additional Diagnostic Analyses}
\label{app:diagnostic_results}

\subsection{Paired Statistics and Horizon Sensitivity}
\label{app:paired_diagnostics}

Table~\ref{tab:app_diagnostic_stats}(a) groups the paired collapse diagnostics
by task: Invariance and Distinguishability differences are shared across the
three horizon rows, while the CF columns report each horizon separately.
Confidence intervals use the paired hierarchical-bootstrap protocol defined in
Appendix~\ref{app:diagnostic_protocol}; negative deltas indicate lower collapse
under PhyLatent. Panel (b) complements these matched diagnostics with held-out
physical--latent rank agreement on fresh random-policy states, without using
the PSG head or the binary collapse definitions.

\paragraph{CF comparison populations.}
The primary CF results use the intersection of LeWM and Full eligibility.
At $H=5$, this contains 13,027, 13,740, 18,948, and 16,093 high--low contrasts
for Cube, TwoRooms, Reacher, and PushT, respectively. Here $N$ counts eligible
contrasts, not episodes, anchors, or action branches. The ablation-family
intersections in Appendix~\ref{app:ablation_success} are smaller.
The Full checkpoints, sampled anchors, and calibrated action branches are the
same across these diagnostic comparisons; their different eligibility sets
explain the different Full CF-H5 values.

\begin{table}[!htb]
\centering
\caption{\textbf{Additional diagnostic statistics.}
Panel (a) combines paired Invariance and Distinguishability differences with
Counterfactual Dynamics (CF) rates across horizons. Invariance and
Distinguishability are reported once per task; only CF varies with $H$.
Rates are percentages and $\Delta=\mathrm{PhyLatent}-\mathrm{LeWM}$ is in
percentage points; brackets give paired 95\% confidence intervals.
Panel (b) reports held-out Spearman rank correlation between physical and
latent pairwise distances (mean $\pm$ standard deviation).}
\label{tab:app_diagnostic_stats}
\label{tab:app_heldout_geometry}
\fontsize{8}{9.2}\selectfont
\renewcommand{\arraystretch}{1.04}
\setlength{\tabcolsep}{1.6pt}
\begin{tabular}{@{}>{\raggedright\arraybackslash}p{.11\linewidth}
>{\centering\arraybackslash}p{\dimexpr.18\linewidth-2\tabcolsep\relax}
>{\centering\arraybackslash}p{\dimexpr.20\linewidth-2\tabcolsep\relax}
>{\centering\arraybackslash}p{\dimexpr.035\linewidth-2\tabcolsep\relax}
>{\raggedleft\arraybackslash}p{\dimexpr.075\linewidth-2\tabcolsep\relax}
>{\raggedleft\arraybackslash}p{\dimexpr.100\linewidth-2\tabcolsep\relax}
>{\raggedright\arraybackslash}p{\dimexpr.300\linewidth-2\tabcolsep\relax}@{}}
\toprule
\multicolumn{7}{c}{\shadecell{groupbg}\textbf{(a) Paired collapse diagnostics and counterfactual horizons}}\\
\shadecell{softgray}\textbf{Task}
& \shadecell{softgray}\shortstack{\textbf{$\Delta$ Inv.}\\\textbf{[95\% CI]}}
& \shadecell{softgray}\shortstack{\textbf{$\Delta$ Dist.}\\\textbf{[95\% CI]}}
& \shadecell{softgray}$\boldsymbol H$
& \shadecell{softgray}\shortstack{\textbf{CF}\\\textbf{LeWM}}
& \shadecell{oursbg}\shortstack{\textbf{CF}\\\textbf{PhyLatent}}
& \shadecell{softgray}\textbf{$\Delta$ CF [95\% CI]}\\
\midrule
& & & 1 & 1.062 & \shadecell{oursbg}0.121 & $-0.942\;[-1.173,\!-0.718]$ \\
& &  & 3 & 5.702 & \shadecell{oursbg}1.769 & $-3.933\;[-4.735,\!-3.059]$ \\
\multirow{-3}{*}{Cube} & \multirow{-3}{*}{\shortstack{$-8.590$\\$[-8.861,\!-8.332]$}} & \multirow{-3}{*}{\shortstack{$-12.684$\\$[-13.109,\!-12.258]$}} & 5 & 14.470 & \shadecell{oursbg}7.646 & $-6.824\;[-8.487,\!-5.251]$ \\
\cdashline{1-7}[0.35pt/1.2pt]
\addlinespace[2pt]
& & & 1 & 24.115 & \shadecell{oursbg}8.474 & $-15.641\;[-17.919,\!-13.247]$ \\
& &  & 3 & 18.284 & \shadecell{oursbg}9.741 & $-8.542\;[-9.818,\!-7.305]$ \\
\multirow{-3}{*}{TwoRooms} & \multirow{-3}{*}{\shortstack{$-12.491$\\$[-12.742,\!-12.257]$}} & \multirow{-3}{*}{\shortstack{$-21.007$\\$[-21.523,\!-20.490]$}} & 5 & 14.105 & \shadecell{oursbg}9.229 & $-4.876\;[-6.307,\!-3.443]$ \\
\cdashline{1-7}[0.35pt/1.2pt]
\addlinespace[2pt]
& & & 1 & 0.767 & \shadecell{oursbg}0.000 & $-0.767\;[-1.019,\!-0.539]$ \\
& &  & 3 & 17.509 & \shadecell{oursbg}14.472 & $-3.037\;[-4.569,\!-1.531]$ \\
\multirow{-3}{*}{Reacher} & \multirow{-3}{*}{\shortstack{$-10.927$\\$[-11.220,\!-10.609]$}} & \multirow{-3}{*}{\shortstack{$-14.144$\\$[-14.518,\!-13.759]$}} & 5 & 15.484 & \shadecell{oursbg}15.104 & \shadecell{warnbg}$-0.380\;[-2.182,\!1.351]$ \\
\cdashline{1-7}[0.35pt/1.2pt]
\addlinespace[2pt]
& & & 1 & 1.517 & \shadecell{oursbg}0.327 & $-1.190\;[-1.459,\!-0.927]$ \\
& &  & 3 & 3.587 & \shadecell{oursbg}1.670 & $-1.918\;[-2.248,\!-1.610]$ \\
\multirow{-3}{*}{PushT} & \multirow{-3}{*}{\shortstack{$-6.474$\\$[-6.621,\!-6.326]$}} & \multirow{-3}{*}{\shortstack{$-7.669$\\$[-8.146,\!-7.189]$}} & 5 & 6.630 & \shadecell{oursbg}2.088 & $-4.542\;[-5.170,\!-3.960]$ \\
\bottomrule
\end{tabular}
\par\vspace{3pt}
\begin{tabular}{@{}>{\raggedright\arraybackslash}p{.20\linewidth}
*{2}{>{\centering\arraybackslash}p{\dimexpr.30\linewidth-2\tabcolsep\relax}}
>{\centering\arraybackslash}p{\dimexpr.20\linewidth-2\tabcolsep\relax}@{}}
\toprule
\multicolumn{4}{c}{\shadecell{groupbg}\textbf{(b) Held-out physical--latent distance rank correlation}}\\
\shadecell{softgray}\textbf{Task} & \shadecell{softgray}\textbf{LeWM} & \shadecell{oursbg}\textbf{PhyLatent} & \shadecell{softgray}$\boldsymbol{\Delta\rho}$\\
\midrule
Cube & $0.1217\pm0.0308$ & \shadecell{oursbg}\textbf{0.1858 $\pm$ 0.0150} & $+0.0641$ \\
TwoRooms & $0.2464\pm0.0733$ & \shadecell{oursbg}\textbf{0.3282 $\pm$ 0.0372} & $+0.0818$ \\
Reacher & $0.0156\pm0.0035$ & \shadecell{oursbg}\textbf{0.1352 $\pm$ 0.0095} & $+0.1196$ \\
PushT & $0.0498\pm0.0192$ & \shadecell{oursbg}\textbf{0.0770 $\pm$ 0.0646} & $+0.0272$ \\
\bottomrule
\end{tabular}
\end{table}

In Table~\ref{tab:app_diagnostic_stats}(a), the CF paired interval at $H{=}5$
excludes zero on Cube, TwoRooms, and PushT, while it includes zero on Reacher. Reacher nevertheless improves at $H{=}1$ and $H{=}3$. The held-out
geometry analysis in panel (b) uses collection seeds 9100--9102, 32 episodes
per seed, and 20,000 shared cross-episode pairs per seed; its absolute
correlations remain modest, so it supports improved rank agreement rather
than an isometric physical-state representation.

\subsection{Diagnostic Sensitivity}
\label{app:diagnostic_sensitivity}

The primary definitions are not the only settings under which the trends can
be observed. We vary appearance severity, physical near/far pairing, and
action-gap strata for the three diagnostics.

\begin{table}[!htb]
\centering
\caption{\textbf{Sensitivity of the collapse diagnostics.}
Appearance severity and physical near/far pairing are shown on the left;
counterfactual action-gap strata are shown on the right.}
\label{tab:app_diagnostic_sensitivity}
\fontsize{8}{9.5}\selectfont
\renewcommand{\arraystretch}{1.00}
\setlength{\tabcolsep}{1.7pt}

\begin{minipage}[t]{0.49\linewidth}
\vspace{0pt}
\centering
\begin{tabular}{@{}lcrrrr@{}}
\toprule
\multicolumn{6}{c}{\shadecell{groupbg}\shortstack[c]{\textbf{(a) Physical Invariance}\\\textbf{Appearance severity}}} \\
\shadecell{softgray}\textbf{Perturbation} & \shadecell{softgray}\textbf{Level} & \shadecell{softgray}$\boldsymbol{N}$
& \shadecell{softgray}\textbf{LeWM} & \shadecell{oursbg}\textbf{PhyLatent}
& \shadecell{softgray}$\boldsymbol{\Delta}$ \\
\midrule
\multirow{3}{*}{Brightness}
& 1 & -- & 2.104 & \shadecell{oursbg}2.097 & $-0.007$ \\
& 2 & -- & 4.830 & \shadecell{oursbg}4.535 & $-0.295$ \\
& 3 & -- & 8.035 & \shadecell{oursbg}7.849 & $-0.186$ \\
\cdashline{1-6}[0.35pt/1.2pt]
\multirow{3}{*}{Tint}
& 1 & -- & 8.364 & \shadecell{oursbg}4.779 & $-3.585$ \\
& 2 & -- & 26.709 & \shadecell{oursbg}10.503 & $-16.205$ \\
& 3 & -- & 36.614 & \shadecell{oursbg}18.560 & $-18.054$ \\
\cdashline{1-6}[0.35pt/1.2pt]
\multirow{3}{*}{Checker}
& 1 & -- & 17.065 & \shadecell{oursbg}6.462 & $-10.603$ \\
& 2 & -- & 35.276 & \shadecell{oursbg}17.161 & $-18.114$ \\
& 3 & -- & 37.238 & \shadecell{oursbg}26.977 & $-10.261$ \\
\addlinespace[3pt]
\multicolumn{6}{c}{\shadecell{groupbg}\shortstack[c]{\textbf{(b) Physical Distinguishability}\\\textbf{Near/far pairing}}} \\
\multicolumn{3}{l}{\shadecell{softgray}\textbf{Physical pair}}
& \shadecell{softgray}\textbf{LeWM} & \shadecell{oursbg}\textbf{PhyLatent}
& \shadecell{softgray}$\boldsymbol{\Delta}$ \\
\midrule
\multicolumn{3}{l}{$Q_1$ vs. $Q_4$} & 45.508 & \shadecell{oursbg}32.824 & $-12.684$ \\
\multicolumn{3}{l}{$Q_1$ vs. $Q_3$} & 45.516 & \shadecell{oursbg}33.342 & $-12.174$ \\
\multicolumn{3}{l}{$Q_2$ vs. $Q_4$} & 49.545 & \shadecell{oursbg}48.015 & $-1.530$ \\
\bottomrule
\end{tabular}
\end{minipage}\hfill
\begin{minipage}[t]{0.49\linewidth}
\vspace{0pt}
\centering
\begin{tabular}{@{}lcrrrr@{}}
\toprule
\multicolumn{6}{c}{\shadecell{groupbg}\shortstack[c]{\textbf{(c) Counterfactual Dynamics}\\\textbf{Action-gap stratification}}} \\
\shadecell{softgray}$\boldsymbol{H}$ & \shadecell{softgray}\textbf{Stratum} & \shadecell{softgray}$\boldsymbol{N}$
& \shadecell{softgray}\textbf{LeWM} & \shadecell{oursbg}\textbf{PhyLatent}
& \shadecell{softgray}$\boldsymbol{\Delta}$ \\
\midrule
\multirow{4}{*}{1}
& Q1 & 1,956 & 2.505 & \shadecell{oursbg}0.051 & $-2.454$ \\
& Q2 & 5,506 & 2.361 & \shadecell{oursbg}0.400 & $-1.961$ \\
& Q3 & 7,946 & 0.617 & \shadecell{oursbg}0.038 & $-0.579$ \\
& Q4 & 6,147 & 0.016 & \shadecell{oursbg}0.000 & $-0.016$ \\
\cdashline{1-6}[0.35pt/1.2pt]
\multirow{4}{*}{3}
& Q1 & 1,908 & 7.495 & \shadecell{oursbg}4.088 & $-3.407$ \\
& Q2 & 2,780 & 6.691 & \shadecell{oursbg}0.935 & $-5.755$ \\
& Q3 & 5,483 & 5.289 & \shadecell{oursbg}1.806 & $-3.483$ \\
& Q4 & 4,806 & 4.890 & \shadecell{oursbg}1.290 & $-3.600$ \\
\cdashline{1-6}[0.35pt/1.2pt]
\multirow{4}{*}{5}
& Q1 & 1,106 & 17.360 & \shadecell{oursbg}13.653 & $-3.707$ \\
& Q2 & 1,201 & 14.488 & \shadecell{oursbg}6.328 & $-8.160$ \\
& Q3 & 6,119 & 14.610 & \shadecell{oursbg}7.583 & $-7.027$ \\
& Q4 & 4,601 & 13.584 & \shadecell{oursbg}6.629 & $-6.955$ \\
\bottomrule
\end{tabular}
\end{minipage}
\end{table}

At $H{=}5$, the reduction is broad across action-gap strata rather than being
concentrated in a single quartile. The near/far sensitivity similarly shows
that the Distinguishability gain is strongest for well-separated physical
states.

\subsection{Frozen Physical-State Probe}
\label{app:probe}

As an auxiliary diagnostic, we freeze the encoder and fit an episode-disjoint
ridge-regression probe using a 70/30 split, $\alpha=1$, at most 12,000 samples
per task, and three split seeds. Because PSG uses related physical targets
during training, the probe is treated as a decodability measure rather than an
independent validation signal.

\begin{table}[!htb]
\centering
\caption{
\textbf{Physical-state probe sensitivity on Cube.}
Lower MSE and mean absolute error (MAE), and higher $R^2$, indicate better physical-state decodability.
}
\label{tab:app_probe_sensitivity}
\footnotesize
\renewcommand{\arraystretch}{0.96}
\setlength{\tabcolsep}{3.2pt}

\begin{tabular}{@{}lrrrrrr@{}}
\toprule
\rowcolor{groupbg}[0pt][0pt]
\multicolumn{7}{c}{\textbf{(a) Probe data efficiency}} \\
\rowcolor{softgray}[0pt][0pt]
\textbf{Training fraction}
& \multicolumn{3}{c}{\textbf{LeWM}}
& \multicolumn{3}{c}{\shadecell{oursbg}\textbf{PhyLatent}} \\
\rowcolor{softgray}[0pt][0pt]
& \textbf{MSE$\downarrow$} & \textbf{MAE$\downarrow$} & \textbf{$R^2\uparrow$}
& \shadecell{oursbg}\textbf{MSE$\downarrow$} & \shadecell{oursbg}\textbf{MAE$\downarrow$} & \shadecell{oursbg}\textbf{$R^2\uparrow$} \\
\midrule
1\%   & 0.498 & 0.571 & 0.268 & \shadecell{oursbg}0.265 & \shadecell{oursbg}0.392 & \shadecell{oursbg}0.610 \\
5\%   & 0.250 & 0.373 & 0.633 & \shadecell{oursbg}0.111 & \shadecell{oursbg}0.209 & \shadecell{oursbg}0.836 \\
10\%  & 0.209 & 0.323 & 0.694 & \shadecell{oursbg}0.099 & \shadecell{oursbg}0.183 & \shadecell{oursbg}0.854 \\
25\%  & 0.186 & 0.295 & 0.727 & \shadecell{oursbg}0.091 & \shadecell{oursbg}0.166 & \shadecell{oursbg}0.866 \\
100\% & 0.174 & 0.280 & 0.744 & \shadecell{oursbg}0.086 & \shadecell{oursbg}0.153 & \shadecell{oursbg}0.874 \\

\addlinespace[3pt]
\rowcolor{groupbg}[0pt][0pt]
\multicolumn{7}{c}{\textbf{(b) Probe target sensitivity}} \\
\rowcolor{softgray}[0pt][0pt]
\textbf{Target group}
& \multicolumn{3}{c}{\textbf{LeWM}}
& \multicolumn{3}{c}{\shadecell{oursbg}\textbf{PhyLatent}} \\
\rowcolor{softgray}[0pt][0pt]
& \textbf{MSE$\downarrow$} & \textbf{MAE$\downarrow$} & \textbf{$R^2\uparrow$}
& \shadecell{oursbg}\textbf{MSE$\downarrow$} & \shadecell{oursbg}\textbf{MAE$\downarrow$} & \shadecell{oursbg}\textbf{$R^2\uparrow$} \\
\midrule
End-effector pose & 0.198 & 0.305 & 0.654 & \shadecell{oursbg}0.010 & \shadecell{oursbg}0.062 & \shadecell{oursbg}0.982 \\
Object pose       & 0.197 & 0.302 & 0.779 & \shadecell{oursbg}0.188 & \shadecell{oursbg}0.277 & \shadecell{oursbg}0.789 \\
Full adapter state& 0.174 & 0.280 & 0.744 & \shadecell{oursbg}0.086 & \shadecell{oursbg}0.153 & \shadecell{oursbg}0.874 \\
\bottomrule
\end{tabular}
\end{table}

The largest target-specific gain occurs for end-effector geometry, whereas
object-pose decodability changes only modestly. This reinforces that state
decodability and planning-effective geometry are related but not identical
properties.



\section{Additional Robustness and Planning Results}
\label{app:robustness_planning}

\subsection{Visual Appearance Stress Test}
\label{app:visual_stress}

Visual-corruption experiments use paired seeds 710--712, 50 episodes per
condition, and CEM 100/top-10. The same corruption family and severity are
applied to both current and goal observations. Let $x\in[0,1]$ denote the
floating-point RGB image, $C(\cdot)$ clipping to $[0,1]$, and $s$ the severity.
The main stress test uses $s=1$. Corruptions are applied before ImageNet
normalization and resizing.

\begin{table}[!htb]
\centering
\caption{
\textbf{Visual-corruption definitions used in the stress test.}
Spatial coordinates $(u,v)$ refer to the pre-resize image.
}
\label{tab:app_visual_definitions}
\small
\renewcommand{\arraystretch}{1.07}
\setlength{\tabcolsep}{3.2pt}

\begin{tabular}{@{}L{0.14\linewidth}L{0.49\linewidth}L{0.27\linewidth}@{}}
\toprule
\shadecell{groupbg}\textbf{Condition}
& \shadecell{groupbg}\textbf{Transform}
& \shadecell{groupbg}\textbf{Sampling / notes} \\
\midrule

Brightness
& $C\!\left((1-0.16s)x+0.07s\right)$
& Deterministic for fixed $s$. \\

Tint
& $C\!\left(x\odot[1+0.16s,\;1-0.12s,\;1+0.05s]\right)$
& Fixed RGB gains. \\

Checker
& $C\!\left(x+s\,q(u,v)[0.08,-0.03,0.06]\right)$,
$q=(\lfloor u/18\rfloor+\lfloor v/18\rfloor)\bmod2$
& Fixed 18-pixel checker pattern and phase. \\

Gaussian noise
& $C(x+0.035s\,\epsilon)$,
$\epsilon\sim\mathcal N(0,I)$
& Resampled on each transform call; separate RNG streams
for current and goal observations. \\

Blur
& Gaussian blur with $\sigma=0.8s$; $5\times5$ kernel for $s\le1$
& Main setting: $5\times5$, $\sigma=0.8$. \\

Background
& $C\!\left(x+s\,r(u)[0.05,-0.02,0.04]\right)$,
\newline $r(u)=\operatorname{linspace}(-1,1,W)_u$
& Deterministic full-image horizontal pattern. \\

Texture
& $C\!\left(x+s\,p(u,v)[0.045,-0.025,0.035]\right)$,
\newline $p=\tfrac12[\sin(u/5)+\cos(v/7)]$
& Deterministic full-image sinusoidal pattern. \\

\bottomrule
\end{tabular}
\end{table}

For Gaussian noise, the current-observation RNG is initialized from the
evaluation seed and the goal-observation RNG from the evaluation seed plus
100000; the generators then advance across transform calls. Background and
texture are full-image additive patterns rather than segmentation-based
background replacement.

\begin{table}[!htb]
\centering
\caption{\textbf{Exact closed-loop results under visual appearance shifts.}
Values are success (\%), mean $\pm$ across-seed standard deviation; $\Delta$
denotes PhyLatent$-$LeWM.}
\label{tab:app_visual_corruption}
\footnotesize
\renewcommand{\arraystretch}{0.96}

\begin{minipage}[t]{0.495\linewidth}
\vspace{0pt}
\centering
\setlength{\tabcolsep}{2.0pt}
\begin{tabular}{@{}lrrr@{}}
\toprule
\multicolumn{4}{c}{\shadecell{groupbg}\textbf{(a) Cube and TwoRooms}} \\
\shadecell{white}\textbf{Condition}
& \shadecell{softgray}\textbf{LeWM}
& \shadecell{oursbg}\textbf{PhyLatent}
& \shadecell{white}\textbf{$\Delta$} \\
\midrule
\multicolumn{4}{l}{\textbf{Cube}} \\
Clean & \shadecell{softgray}$72.67\pm6.43$ & \shadecell{oursbg}$78.67\pm5.77$ & $+6.00$ \\
Brightness & \shadecell{softgray}$71.33\pm7.57$ & \shadecell{oursbg}$71.33\pm6.43$ & $0.00$ \\
Tint & \shadecell{softgray}$64.00\pm8.00$ & \shadecell{oursbg}$74.00\pm8.72$ & $+10.00$ \\
Checker & \shadecell{softgray}$54.00\pm3.46$ & \shadecell{oursbg}$64.00\pm2.00$ & $+10.00$ \\
Noise & \shadecell{softgray}$72.67\pm6.11$ & \shadecell{oursbg}$79.33\pm5.03$ & $+6.67$ \\
Blur & \shadecell{softgray}$74.00\pm3.46$ & \shadecell{oursbg}$79.33\pm3.06$ & $+5.33$ \\
Background & \shadecell{softgray}$54.67\pm2.31$ & \shadecell{oursbg}$72.67\pm6.11$ & $+18.00$ \\
Texture & \shadecell{softgray}$58.00\pm0.00$ & \shadecell{oursbg}$78.00\pm3.46$ & $+20.00$ \\
\cdashline{1-4}[0.35pt/1.2pt]
\multicolumn{4}{l}{\textbf{TwoRooms}} \\
Clean & \shadecell{softgray}$85.33\pm3.06$ & \shadecell{oursbg}$95.33\pm1.15$ & $+10.00$ \\
Brightness & \shadecell{softgray}$80.67\pm6.43$ & \shadecell{oursbg}$90.00\pm6.93$ & $+9.33$ \\
Tint & \shadecell{softgray}$80.67\pm5.03$ & \shadecell{oursbg}$92.67\pm2.31$ & $+12.00$ \\
Checker & \shadecell{softgray}$80.00\pm5.29$ & \shadecell{oursbg}$92.00\pm4.00$ & $+12.00$ \\
Noise & \shadecell{softgray}$84.00\pm2.00$ & \shadecell{oursbg}$87.33\pm2.31$ & $+3.33$ \\
Blur & \shadecell{softgray}$60.00\pm7.21$ & \shadecell{oursbg}$60.67\pm4.62$ & $+0.67$ \\
Background & \shadecell{softgray}$47.33\pm9.24$ & \shadecell{oursbg}$22.67\pm1.15$ & $-24.67$ \\
Texture & \shadecell{softgray}$77.33\pm1.15$ & \shadecell{oursbg}$44.67\pm2.31$ & $-32.67$ \\
\bottomrule
\end{tabular}
\end{minipage}
\hfill
\begin{minipage}[t]{0.495\linewidth}
\vspace{0pt}
\centering
\setlength{\tabcolsep}{2.0pt}
\begin{tabular}{@{}lrrr@{}}
\toprule
\multicolumn{4}{c}{\shadecell{groupbg}\textbf{(b) Reacher and PushT}} \\
\shadecell{white}\textbf{Condition}
& \shadecell{softgray}\textbf{LeWM}
& \shadecell{oursbg}\textbf{PhyLatent}
& \shadecell{white}\textbf{$\Delta$} \\
\midrule
\multicolumn{4}{l}{\textbf{Reacher}} \\
Clean & \shadecell{softgray}$80.67\pm1.15$ & \shadecell{oursbg}$74.67\pm7.57$ & $-6.00$ \\
Brightness & \shadecell{softgray}$78.67\pm2.31$ & \shadecell{oursbg}$78.00\pm5.29$ & $-0.67$ \\
Tint & \shadecell{softgray}$70.00\pm9.17$ & \shadecell{oursbg}$72.67\pm6.11$ & $+2.67$ \\
Checker & \shadecell{softgray}$68.67\pm4.16$ & \shadecell{oursbg}$76.00\pm5.29$ & $+7.33$ \\
Noise & \shadecell{softgray}$80.00\pm3.46$ & \shadecell{oursbg}$78.00\pm3.46$ & $-2.00$ \\
Blur & \shadecell{softgray}$76.00\pm12.49$ & \shadecell{oursbg}$76.00\pm7.21$ & $0.00$ \\
Background & \shadecell{softgray}$63.33\pm2.31$ & \shadecell{oursbg}$28.00\pm6.00$ & $-35.33$ \\
Texture & \shadecell{softgray}$82.67\pm1.15$ & \shadecell{oursbg}$44.67\pm10.26$ & $-38.00$ \\
\cdashline{1-4}[0.35pt/1.2pt]
\multicolumn{4}{l}{\textbf{PushT}} \\
Clean & \shadecell{softgray}$86.00\pm2.00$ & \shadecell{oursbg}$84.00\pm5.29$ & $-2.00$ \\
Brightness & \shadecell{softgray}$83.33\pm3.06$ & \shadecell{oursbg}$79.33\pm7.57$ & $-4.00$ \\
Tint & \shadecell{softgray}$38.00\pm3.46$ & \shadecell{oursbg}$71.33\pm9.02$ & $+33.33$ \\
Checker & \shadecell{softgray}$81.33\pm5.03$ & \shadecell{oursbg}$81.33\pm10.26$ & $0.00$ \\
Noise & \shadecell{softgray}$62.67\pm3.06$ & \shadecell{oursbg}$68.67\pm3.06$ & $+6.00$ \\
Blur & \shadecell{softgray}$79.33\pm4.16$ & \shadecell{oursbg}$81.33\pm5.77$ & $+2.00$ \\
Background & \shadecell{softgray}$78.67\pm7.02$ & \shadecell{oursbg}$77.33\pm6.43$ & $-1.33$ \\
Texture & \shadecell{softgray}$80.00\pm3.46$ & \shadecell{oursbg}$81.33\pm3.06$ & $+1.33$ \\
\bottomrule
\end{tabular}
\end{minipage}
\end{table}

The condition-wise values clarify the mechanism discussed in the main text:
Cube exhibits broad transfer beyond the simple appearance perturbations used by
SVIP, whereas scene-level background and texture changes expose clear failure
modes on TwoRooms and Reacher.

\subsection{Goal-Separation Stress Test}
\label{app:goal_offset_stress}

Each temporal-offset point uses paired seeds 730--732, 50 episodes per seed,
and CEM 100/top-10. Larger offset denotes greater temporal separation, not a
strictly monotonic difficulty scale.

\begin{table}[!htb]
\centering
\caption{\textbf{Exact closed-loop success (\%) under increasing temporal goal offsets.}
Avg. is the unweighted mean across the five offsets.}
\label{tab:app_goal_offsets}
\small
\renewcommand{\arraystretch}{1.03}
\setlength{\tabcolsep}{2.7pt}
\begin{tabular}{@{}llrrrrrr@{}}
\toprule
\shadecell{groupbg}\textbf{Task}
& \shadecell{groupbg}\textbf{Method}
& \shadecell{groupbg}\textbf{10}
& \shadecell{groupbg}\textbf{25}
& \shadecell{groupbg}\textbf{40}
& \shadecell{groupbg}\textbf{50}
& \shadecell{groupbg}\textbf{75}
& \shadecell{groupbg}\textbf{Avg.} \\
\midrule
\multirow{2}{*}{Cube}
& \shadecell{softgray}LeWM & \shadecell{softgray}91.33 & \shadecell{softgray}54.67 & \shadecell{softgray}43.33 & \shadecell{softgray}46.00 & \shadecell{softgray}60.67 & \shadecell{softgray}59.20 \\
& \shadecell{oursbg}\textbf{PhyLatent} & \shadecell{oursbg}92.67 & \shadecell{oursbg}59.33 & \shadecell{oursbg}48.00 & \shadecell{oursbg}46.67 & \shadecell{oursbg}68.00 & \shadecell{oursbg}\textbf{62.93} \\
\cdashline{1-8}[0.40pt/1.4pt]
\multirow{2}{*}{TwoRooms}
& \shadecell{softgray}LeWM & \shadecell{softgray}98.00 & \shadecell{softgray}86.67 & \shadecell{softgray}66.00 & \shadecell{softgray}49.33 & \shadecell{softgray}16.00 & \shadecell{softgray}63.20 \\
& \shadecell{oursbg}\textbf{PhyLatent} & \shadecell{oursbg}100.00 & \shadecell{oursbg}98.00 & \shadecell{oursbg}80.67 & \shadecell{oursbg}69.33 & \shadecell{oursbg}46.67 & \shadecell{oursbg}\textbf{78.93} \\
\cdashline{1-8}[0.40pt/1.4pt]
\multirow{2}{*}{Reacher}
& \shadecell{softgray}LeWM & \shadecell{softgray}87.33 & \shadecell{softgray}79.33 & \shadecell{softgray}74.67 & \shadecell{softgray}78.00 & \shadecell{softgray}65.33 & \shadecell{softgray}\textbf{76.93} \\
& \shadecell{oursbg}\textbf{PhyLatent} & \shadecell{oursbg}81.33 & \shadecell{oursbg}74.67 & \shadecell{oursbg}73.33 & \shadecell{oursbg}75.33 & \shadecell{oursbg}66.00 & \shadecell{oursbg}74.13 \\
\cdashline{1-8}[0.40pt/1.4pt]
\multirow{2}{*}{PushT}
& \shadecell{softgray}LeWM & \shadecell{softgray}90.67 & \shadecell{softgray}80.67 & \shadecell{softgray}57.33 & \shadecell{softgray}35.33 & \shadecell{softgray}26.00 & \shadecell{softgray}58.00 \\
& \shadecell{oursbg}\textbf{PhyLatent} & \shadecell{oursbg}88.67 & \shadecell{oursbg}82.67 & \shadecell{oursbg}64.67 & \shadecell{oursbg}44.67 & \shadecell{oursbg}30.00 & \shadecell{oursbg}\textbf{62.14} \\
\bottomrule
\end{tabular}
\end{table}

For the Cube physical-displacement stratification, the difficulty variable is
the true 3D start--goal cube displacement
\begin{equation}
D_i
=
\left\|
p^{\mathrm{cube}}_{t+25}
-
p^{\mathrm{cube}}_{t}
\right\|_2 .
\label{eq:app_cube_displacement}
\end{equation}
The coordinates are raw simulator world coordinates in meters and are converted
to centimeters only for display. The stratification uses seeds 730--732 with
100 paired instances per seed, goal offset 25, and CEM 100/top-10 with 10
iterations. Quartile boundaries are computed once from the shared displacement
set and then applied identically to both models.

\begin{table}[!htb]
\centering
\caption{\textbf{Cube physical-displacement stratification.}
The 300 paired instances are split using shared quartile boundaries computed
once from the true start--goal cube displacement; each quartile contains 75
instances. The displayed ranges are Q1: $<0.039$ cm, Q2: $0.039$--$9.101$ cm,
Q3: $9.101$--$19.966$ cm, and Q4: $19.966$--$49.888$ cm.}
\label{tab:app_cube_displacement}
\small
\renewcommand{\arraystretch}{1.05}
\setlength{\tabcolsep}{3.0pt}
\begin{tabular}{@{}lccccc@{}}
\toprule
\shadecell{groupbg}\textbf{Method}
& \shadecell{groupbg}\textbf{Q1}
& \shadecell{groupbg}\textbf{Q2}
& \shadecell{groupbg}\textbf{Q3}
& \shadecell{groupbg}\textbf{Q4}
& \shadecell{groupbg}\makecell[c]{\textbf{Q1$\to$Q4}\\\textbf{Rel. Drop$\downarrow$}} \\
\midrule
\shadecell{softgray}LeWM & \shadecell{softgray}100.00 & \shadecell{softgray}81.33 & \shadecell{softgray}48.00 & \shadecell{softgray}26.67 & \shadecell{softgray}73.33 \\
\shadecell{oursbg}\textbf{PhyLatent} & \shadecell{oursbg}100.00 & \shadecell{oursbg}88.00 & \shadecell{oursbg}52.00 & \shadecell{oursbg}\textbf{60.00} & \shadecell{oursbg}\textbf{40.00} \\
\bottomrule
\end{tabular}
\end{table}

The temporal and physical views agree on the main qualitative pattern: the
largest benefits occur when planning must preserve useful state relations over
larger changes, while Reacher remains the principal exception.

\subsection{Latent and Action Perturbation Robustness}
\label{app:latent_action_robustness}

We use paired seeds 720--722, 50 instances per seed and model, goal offset 25,
and a 50-step execution budget. Evaluation uses 50 parallel environments.
CEM uses 100 candidates, 10 elites, and 10 iterations; prediction and receding
horizons are five blocks, with five low-level actions per block.

For latent perturbation, $\hat z'=\hat z+\sigma_z\epsilon$ with
$\epsilon\sim\mathcal N(0,I)$ and
$\sigma_z\in\{0,0.02,0.05,0.10,0.20\}$.
Noise is added elementwise to the entire predicted embedding tensor after
every predictor call during imagined rollouts, without subsequent
normalization or clipping. Observation and goal encodings are not perturbed.
A CUDA generator initialized from the evaluation seed advances across calls,
without resetting at episode boundaries.

For action perturbation,
$a'=\operatorname{clip}(a+\lambda(a_{\max}-a_{\min})\epsilon,
a_{\min},a_{\max})$, with
$\lambda\in\{0,0.01,0.02,0.05,0.10\}$.
Noise is injected into the policy output immediately before execution, after
CEM has selected the action. A NumPy generator initialized from the evaluation
seed draws a fresh Gaussian tensor of the action shape on each policy call.
For non-finite action ranges the span defaults to 2; only finite action
elements are modified. Both models use matched seeded noise streams; this
does not enforce identical elementwise draws after differing call sequences.
The zero-noise control is shared between the two sweeps.

\begin{table}[!htb]
\centering
\caption{\textbf{Cube planning robustness to latent and action perturbations.}
Values are success (\%), mean $\pm$ across-seed standard deviation; $\Delta$
denotes PhyLatent$-$LeWM.}
\label{tab:app_latent_action_robustness}
\small
\renewcommand{\arraystretch}{1.03}

\begin{minipage}[t]{0.49\linewidth}
\centering
\setlength{\tabcolsep}{2.0pt}
\begin{tabular}{@{}cccc@{}}
\toprule
\multicolumn{4}{c}{\shadecell{groupbg}\textbf{(a) Latent perturbation}} \\
\shadecell{white}\textbf{$\sigma_z$}
& \shadecell{softgray}\textbf{LeWM}
& \shadecell{oursbg}\textbf{PhyLatent}
& \shadecell{white}\textbf{$\Delta$} \\
\midrule
0.00 & \shadecell{softgray}58.67$\pm$8.33 & \shadecell{oursbg}69.33$\pm$5.03 & +10.67 \\
0.02 & \shadecell{softgray}60.00$\pm$9.17 & \shadecell{oursbg}68.00$\pm$5.29 & +8.00 \\
0.05 & \shadecell{softgray}62.00$\pm$9.17 & \shadecell{oursbg}72.67$\pm$11.02 & +10.67 \\
0.10 & \shadecell{softgray}60.67$\pm$3.06 & \shadecell{oursbg}67.33$\pm$1.15 & +6.67 \\
0.20 & \shadecell{softgray}58.67$\pm$8.33 & \shadecell{oursbg}70.00$\pm$2.00 & +11.33 \\
\bottomrule
\end{tabular}
\end{minipage}
\hfill
\begin{minipage}[t]{0.49\linewidth}
\centering
\setlength{\tabcolsep}{2.0pt}
\begin{tabular}{@{}cccc@{}}
\toprule
\multicolumn{4}{c}{\shadecell{groupbg}\textbf{(b) Action perturbation}} \\
\shadecell{white}\textbf{$\lambda$}
& \shadecell{softgray}\textbf{LeWM}
& \shadecell{oursbg}\textbf{PhyLatent}
& \shadecell{white}\textbf{$\Delta$} \\
\midrule
0.00 & \shadecell{softgray}58.67$\pm$8.33 & \shadecell{oursbg}69.33$\pm$5.03 & +10.67 \\
0.01 & \shadecell{softgray}55.33$\pm$4.16 & \shadecell{oursbg}65.33$\pm$5.03 & +10.00 \\
0.02 & \shadecell{softgray}54.67$\pm$6.11 & \shadecell{oursbg}67.33$\pm$3.06 & +12.67 \\
0.05 & \shadecell{softgray}57.33$\pm$6.11 & \shadecell{oursbg}62.00$\pm$5.29 & +4.67 \\
0.10 & \shadecell{softgray}54.67$\pm$5.03 & \shadecell{oursbg}56.67$\pm$3.06 & +2.00 \\
\bottomrule
\end{tabular}
\end{minipage}
\end{table}

PhyLatent remains higher at every tested latent- and action-noise level. The
advantage narrows at the strongest action perturbations, where low-level
execution increasingly dominates the error budget.

\subsection{Planner-Budget Sensitivity}
\label{app:planner_budget}

We use paired seeds 750--752, 50 instances per seed and model, and batches of
five environments. Candidate/elite counts are 100/10, 300/30, and 600/60,
preserving a 10\% elite fraction with ten CEM iterations. Goal offset is 25,
execution budget is 50 environment steps, and prediction/receding horizons
and action-block size remain 5/5/5.

Runtime is evaluation wall-clock seconds per episode: elapsed time divided
by the number of episodes for each seed, then averaged across seeds.
Timing starts after model, dataset, world, and solver initialization and
includes policy setup, planning, environment stepping/rendering, the Python
evaluation loop, cache clearing, and result-metadata preparation (including
weight hashing and package queries). Warm-up time is not separately removed.
These values do not isolate planner inference latency. 

\begin{table}[!htb]
\centering
\caption{\textbf{Cube CEM search-budget sensitivity.}
Runtime is evaluation wall-clock seconds per episode on the same machine,
averaged across seeds and intended only as a relative reference.}
\label{tab:app_planner_budget}
\small
\renewcommand{\arraystretch}{1.05}
\setlength{\tabcolsep}{3.0pt}
\begin{tabular}{@{}crrrrrr@{}}
\toprule
\shadecell{white}\multirow{2}{*}{\textbf{CEM samples}}
& \multicolumn{3}{c}{\shadecell{groupbg}\textbf{Success (\%)}}
& \multicolumn{3}{c}{\shadecell{groupbg}\textbf{Runtime}} \\
& \shadecell{softgray}\textbf{LeWM}
& \shadecell{oursbg}\textbf{PhyLatent}
& \shadecell{white}\textbf{$\Delta$}
& \shadecell{softgray}\textbf{LeWM s/ep}
& \shadecell{oursbg}\textbf{Phy s/ep}
& \shadecell{white}\textbf{$\Delta t$} \\
\midrule
100 & \shadecell{softgray}64.00 & \shadecell{oursbg}75.33 & +11.33 & \shadecell{softgray}0.863 & \shadecell{oursbg}0.949 & +10.0\% \\
\shadecell{warnbg}\textbf{300} & \shadecell{softgray}70.67 & \shadecell{oursbg}82.00 & +11.33 & \shadecell{softgray}0.891 & \shadecell{oursbg}0.996 & +11.9\% \\
600 & \shadecell{softgray}72.00 & \shadecell{oursbg}80.00 & +8.00 & \shadecell{softgray}1.044 & \shadecell{oursbg}1.142 & +9.4\% \\
\bottomrule
\end{tabular}
\end{table}

The planning advantage is preserved across all three search budgets, indicating
that the main result is not specific to the default CEM sample count.

\subsection{Qualitative Rollout Reproduction}
\label{app:qualitative_protocol}

Figure~\ref{fig:qualitative_collapse} presents independently executed LeWM and
PhyLatent closed-loop rollouts from matched Cube states. These differ from the
LeWM alternative-candidate replays in Appendix~\ref{app:decision_cases}.
The selected instances are as follows:
\begin{itemize}[leftmargin=1.2em,itemsep=2pt,topsep=2pt]
\item \textbf{Invariance:} dataset row 20164 (episode 100, start step 64),
goal row 20221, offset 57, and reset seed 24264. The shared first seven
snapshot times are $0,3,5,8,11,14,16$; LeWM/PhyLatent terminate at 50/19.
\item \textbf{Distinguishability:} row 1201110 (episode 5975, start step 135),
goal row 1201135, offset 25, and reset seed 1205210. Snapshot times are
$0,3,6,9,11,14,17$; termination times are 50/20.
\item \textbf{Counterfactual:} row 891917 (episode 4437, start step 80),
goal row 891960, offset 43, and reset seed 896017. Snapshot times are
$0,3,7,10,14,17,21$; termination times are 50/24.
\end{itemize}
All three use planner seed 600, one environment, CEM 300/30/10
(candidates/elites/iterations), prediction/receding horizons of five blocks,
five actions per block, and a 50-step budget. The models share the initial
physical state, goal, and first CEM candidate pool, then execute independent
closed-loop plans.

The first seven snapshots use the first seven entries of
$\operatorname{round}(\operatorname{linspace}(0,T_{\rm Phy},8))$;
the last snapshot is each model's actual terminal state. Distances are 3D cube--goal Euclidean errors in
centimeters; boxes mark terminal outcomes, not a diagnosed collapse time.
The invariance case applies checker corruption to current and goal inputs
throughout planning. Only its $t=0$ display shows the corruption; subsequent
clean simulator renders make the physical consequences visible. The other
two cases use clean inputs. All panels use the original camera cropped to
$224\times176$. These selected examples illustrate behavior and are not used
to estimate population success rates.


\section{Full Ablation Results}
\label{app:ablation_full}

\subsection{Cross-Task Pathway and Component Ablations}
\label{app:ablation_success}

Table~\ref{tab:app_full_cross_task_ablation} places all four environments in
task blocks, with metrics in rows and ablation variants in columns. Full
denotes PhyLatent; each other column removes the named pathway or component.
Success is reported as mean $\pm$ across-seed standard deviation; Inv.,
Dist., and CF-H5 are the corresponding collapse diagnostics.
For CF-H5, each task uses a fixed eligibility intersection across its comparison family:
ten checkpoints for Cube and nine for each remaining task, including LeWM
and Full. The resulting $H=5$ contrast counts are 10,162, 12,036, 17,377,
and 15,380 for Cube, TwoRooms, Reacher, and PushT. These differ from the
pairwise populations in Table~\ref{tab:app_diagnostic_stats}, accounting for
the different Full CF-H5 values without changing checkpoints or sampled
anchors/actions. Lower diagnostic values are better.

The Cube intersection includes eight ablation checkpoints, one more than the
seven removal columns displayed. Its source identifiers are
\texttt{wo\_physical\_group}, \texttt{wo\_invariance\_group},
\texttt{wo\_counterfactual\_group}, \texttt{wo\_psg}, \texttt{wo\_fra},
\texttt{wo\_svic}, \texttt{wo\_casc}, and \texttt{wo\_ld}.
The displayed SVIP removal uses \texttt{wo\_invariance\_group}, whose
configuration disables only SVIP; \texttt{wo\_svic} also participates in the
intersection. The other task families comprise five single-component and
two group removals in addition to LeWM and Full.

\begin{table}[!htb]
\centering
\caption{\textbf{Full cross-task pathway and component ablations.}
Each task block reports closed-loop success and the three primary collapse
diagnostics, with CF-H5 evaluated at the planner-aligned horizon.
Metrics are arranged in rows and variants in columns; every column except
Full removes the named group or component.}
\label{tab:app_full_cross_task_ablation}
\fontsize{8}{9.2}\selectfont
\renewcommand{\arraystretch}{1.00}
\setlength{\tabcolsep}{1.4pt}
\begin{tabular}{@{}>{\raggedright\arraybackslash}p{.13\linewidth}
>{\columncolor{oursbg}[0pt][0pt]\raggedleft\arraybackslash}p{\dimexpr.10875\linewidth-2\tabcolsep\relax}
>{\columncolor{softgray}[0pt][0pt]\raggedleft\arraybackslash}p{\dimexpr.10875\linewidth-2\tabcolsep\relax}
*{2}{>{\raggedleft\arraybackslash}p{\dimexpr.10875\linewidth-2\tabcolsep\relax}}
>{\columncolor{softgray}[0pt][0pt]\raggedleft\arraybackslash}p{\dimexpr.10875\linewidth-2\tabcolsep\relax}
>{\columncolor{softgray}[0pt][0pt]\raggedleft\arraybackslash}p{\dimexpr.10875\linewidth-2\tabcolsep\relax}
*{2}{>{\raggedleft\arraybackslash}p{\dimexpr.10875\linewidth-2\tabcolsep\relax}}@{}}
\toprule
\shadecell{softgray}\textbf{Metric} & \shadecell{oursbg}\textbf{Full} & \shadecell{softgray}\shortstack{\textbf{Ground.}\\\textbf{grp.}}
& \shadecell{softgray}\textbf{PSG} & \shadecell{softgray}\textbf{FRA} & \shadecell{softgray}\textbf{SVIP}
& \shadecell{softgray}\shortstack{\textbf{CF}\\\textbf{grp.}} & \shadecell{softgray}\textbf{CASP} & \shadecell{softgray}\textbf{LD} \\
\midrule
\multicolumn{9}{l}{\textbf{Cube}} \\
Success $\uparrow$
& \textbf{81.67$\pm$3.01} & 76.50$\pm$3.15 & 77.00$\pm$3.63 & 79.33$\pm$1.51
& 80.67$\pm$4.27 & 81.17$\pm$2.23 & 79.00$\pm$0.89 & 79.50$\pm$1.97 \\
Inv. $\downarrow$ & 10.99 & 11.85 & 13.44 & 8.46 & 9.67 & 9.74 & 9.46 & 6.93 \\
Dist. $\downarrow$ & 32.82 & 40.11 & 39.33 & 28.85 & 29.71 & 27.26 & 28.42 & 28.81 \\
CF-H5 $\downarrow$ & 5.934 & 13.629 & 13.836 & 7.567 & 8.611 & 5.737 & 6.908 & 6.288 \\
\midrule
\multicolumn{9}{l}{\textbf{TwoRooms}} \\
Success $\uparrow$
& \textbf{96.83$\pm$2.14} & 98.00$\pm$1.26 & 97.50$\pm$0.84 & 96.33$\pm$1.63
& 95.50$\pm$1.52 & 98.00$\pm$1.41 & 97.83$\pm$1.17 & 96.50$\pm$1.97 \\
Inv. $\downarrow$ & 4.69 & 3.61 & 14.55 & 15.91 & 11.77 & 4.11 & 5.95 & 13.61 \\
Dist. $\downarrow$ & 26.67 & 34.90 & 37.33 & 37.15 & 37.53 & 35.99 & 37.38 & 37.85 \\
CF-H5 $\downarrow$ & 4.894 & 5.284 & 3.473 & 5.118 & 4.038 & 2.684 & 4.595 & 5.251 \\
\midrule
\multicolumn{9}{l}{\textbf{Reacher}} \\
Success $\uparrow$
& \textbf{82.17$\pm$3.60} & 65.00$\pm$6.32 & 51.00$\pm$5.59 & 33.50$\pm$2.17
& 63.17$\pm$4.62 & 69.17$\pm$3.71 & 56.67$\pm$5.13 & 79.83$\pm$3.37 \\
Inv. $\downarrow$ & 7.66 & 16.08 & 16.73 & 18.17 & 17.05 & 17.60 & 15.01 & 15.32 \\
Dist. $\downarrow$ & 33.46 & 33.43 & 26.96 & 27.78 & 24.36 & 42.52 & 44.50 & 41.41 \\
CF-H5 $\downarrow$ & 15.020 & 14.657 & 17.702 & 22.823 & 21.252 & 19.163 & 18.415 & 19.808 \\
\midrule
\multicolumn{9}{l}{\textbf{PushT}} \\
Success $\uparrow$
& \textbf{85.50$\pm$3.62} & 59.83$\pm$3.31 & 63.67$\pm$5.05 & 64.17$\pm$1.94
& 76.67$\pm$2.94 & 78.50$\pm$5.89 & 84.33$\pm$1.37 & 72.83$\pm$5.04 \\
Inv. $\downarrow$ & 3.38 & 5.73 & 5.34 & 4.33 & 4.58 & 3.86 & 3.71 & 4.15 \\
Dist. $\downarrow$ & 30.71 & 32.21 & 34.82 & 29.18 & 29.60 & 28.43 & 29.58 & 28.54 \\
CF-H5 $\downarrow$ & 1.925 & 3.212 & 3.524 & 2.107 & 2.133 & 2.087 & 2.074 & 1.925 \\
\bottomrule
\end{tabular}
\end{table}

\subsection{Cube CF-H5 Paired Differences}
\label{app:cube_ablation_diagnostics}

The Cube counterfactual ablations are additionally compared on the shared
common-eligibility set. Table~\ref{tab:app_cube_ablation_ci} reports the paired
change relative to Full; the remaining cross-task interpretation follows the
same non-orthogonal pattern discussed in the main text.

The paired bootstrap uses the seed--anchor procedure in
Appendix~\ref{app:diagnostic_protocol}, recomputing the difference of pooled
failure rates on the fixed family intersection. RNG seeds 151502--151509
follow the eight source identifiers listed above in order; the displayed
SVIP-removal interval therefore uses 151503.

\begin{table}[!htb]
\centering
\caption{\textbf{Cube component-level CF-H5 differences on the shared
common-eligibility set.} Columns show Full PhyLatent and removals of each
named group or component. The paired change relative to Full and the lower
and upper bounds of its 95\% confidence interval are shown in separate rows.}
\label{tab:app_cube_ablation_ci}
\fontsize{8}{9.2}\selectfont
\renewcommand{\arraystretch}{1.08}
\setlength{\tabcolsep}{2.0pt}
\begin{tabular}{@{}>{\raggedright\arraybackslash}p{.19\linewidth}
>{\columncolor{oursbg}[0pt][0pt]\raggedleft\arraybackslash}p{\dimexpr.10125\linewidth-2\tabcolsep\relax}
>{\columncolor{softgray}[0pt][0pt]\raggedleft\arraybackslash}p{\dimexpr.10125\linewidth-2\tabcolsep\relax}
*{2}{>{\raggedleft\arraybackslash}p{\dimexpr.10125\linewidth-2\tabcolsep\relax}}
>{\columncolor{softgray}[0pt][0pt]\raggedleft\arraybackslash}p{\dimexpr.10125\linewidth-2\tabcolsep\relax}
>{\columncolor{softgray}[0pt][0pt]\raggedleft\arraybackslash}p{\dimexpr.10125\linewidth-2\tabcolsep\relax}
*{2}{>{\raggedleft\arraybackslash}p{\dimexpr.10125\linewidth-2\tabcolsep\relax}}@{}}
\toprule
\shadecell{softgray}\textbf{Metric} & \shadecell{oursbg}\textbf{Full} & \shadecell{softgray}\shortstack{\textbf{Ground.}\\\textbf{grp.}}
& \shadecell{softgray}\textbf{PSG} & \shadecell{softgray}\textbf{FRA} & \shadecell{softgray}\textbf{SVIP}
& \shadecell{softgray}\shortstack{\textbf{CF}\\\textbf{grp.}} & \shadecell{softgray}\textbf{CASP} & \shadecell{softgray}\textbf{LD} \\
\midrule
$\Delta$CF & -- & $+7.695$ & $+7.902$ & $+1.634$ & $+2.677$ & $-0.197$ & $+0.974$ & $+0.354$ \\
95\% CI lower & & $+6.100$ & $+5.732$ & $+0.799$ & $+1.763$ & $-0.958$ & $+0.010$ & $-0.467$ \\
95\% CI upper & \multirow{-2}{*}{reference} & $+9.347$ & $+10.057$ & $+2.476$ & $+3.583$ & $+0.576$ & $+2.029$ & $+1.137$ \\
\bottomrule
\end{tabular}
\end{table}

Across tasks, component effects remain strongly non-orthogonal: variants that
improve one collapse statistic can still degrade success or another part of
the latent geometry, consistent with the three pathways jointly shaping a
shared representation rather than isolated one-to-one corrections.


\end{document}